%% file: iclr2027_conference.tex
\documentclass{article}
\usepackage{iclr2027_conference,times}

\input{math_commands}

\usepackage{hyperref}
\usepackage{url}
\usepackage{booktabs}
\usepackage{amsmath}
\usepackage{multirow}
\usepackage{enumitem}
\usepackage{graphicx}
\usepackage{wrapfig}
\usepackage{amssymb}
\usepackage{algorithm}
\usepackage{algpseudocode}
\usepackage[most]{tcolorbox}  
\usepackage[dvipsnames]{xcolor}

\definecolor{OurColor}{RGB}{235, 238, 242}
\usepackage[table]{xcolor}

\usepackage[most]{tcolorbox}
\usepackage{tikz}
\usepackage{tabularx}
\usepackage{enumitem}
\usetikzlibrary{arrows.meta,positioning,fit,calc}

\definecolor{CaseBlue}{RGB}{42,92,145}
\definecolor{CaseBlueBg}{RGB}{241,247,253}
\definecolor{CaseRed}{RGB}{166,55,55}
\definecolor{CaseRedBg}{RGB}{253,243,243}
\definecolor{CaseGreen}{RGB}{35,119,82}
\definecolor{CaseGreenBg}{RGB}{239,249,244}
\definecolor{CaseGold}{RGB}{161,112,21}
\definecolor{CaseGoldBg}{RGB}{255,249,235}
\definecolor{CaseGray}{RGB}{90,98,108}
\definecolor{CaseGrayBg}{RGB}{247,248,250}

\newtcolorbox{caseoverview}[1]{
  enhanced, breakable,
  colback=CaseBlueBg, colframe=CaseBlue,
  boxrule=0.8pt, arc=1.5mm,
  left=1.5mm,right=1.5mm,top=1mm,bottom=1mm,
  fonttitle=\bfseries, title={#1}
}
\newtcolorbox{failurebox}[1]{
  enhanced, breakable,
  colback=CaseRedBg, colframe=CaseRed,
  boxrule=0.8pt, arc=1.5mm,
  left=1.5mm,right=1.5mm,top=1mm,bottom=1mm,
  fonttitle=\bfseries, title={#1}
}
\newtcolorbox{successbox}[1]{
  enhanced, breakable,
  colback=CaseGreenBg, colframe=CaseGreen,
  boxrule=0.8pt, arc=1.5mm,
  left=1.5mm,right=1.5mm,top=1mm,bottom=1mm,
  fonttitle=\bfseries, title={#1}
}
\newtcolorbox{evidencebox}[1]{
  enhanced, breakable,
  colback=CaseGoldBg, colframe=CaseGold,
  boxrule=0.8pt, arc=1.5mm,
  left=1.5mm,right=1.5mm,top=1mm,bottom=1mm,
  fonttitle=\bfseries, title={#1}
}
\newtcolorbox{tracebox}[1]{
  enhanced, breakable,
  colback=CaseGrayBg, colframe=CaseGray,
  boxrule=0.6pt, arc=1mm,
  left=1.5mm,right=1.5mm,top=1mm,bottom=1mm,
  fonttitle=\bfseries, title={#1}
}

\usepackage{hyperref}      
\usepackage{fontawesome5}  
\usepackage{twemojis}      
\usepackage{cuted} 

\title{MAS-Gate: Unlocking Multi-Agent System Lifecycle via Node-Centric Quantification}

\author{
\bf Hehai Lin$^{1}$\thanks{Equal contribution}~, 
Zixuan Wang$^{1}$\footnotemark[1]~, 
Ming Yan$^{1}$\footnotemark[1]~, 
Peidong Wang$^{1}$, 
Shilei Cao$^{2}$, \\
\bf  ~Minzhi Li$^{4,5}$, 
Bosheng Ding$^{3}$, 
Juepeng Zheng$^{2}$, 
Chengwei Qin{$^{1}$}\thanks{Corresponding to \href{mailto:chengweiqin@hkust-gz.edu.cn}{chengweiqin@hkust-gz.edu.cn}}
\\
  $^{1}$The Hong Kong University of Science and Technology (Guangzhou), \\
  $^{2}$Sun Yat-sen University, $^{3}$Nanyang Technological University, \\
  $^{4}$National University of Singapore,
  $^{5}$Institute for Infocomm Research (I$^2$R), A*STAR
}

\iclrfinalcopy
\begin{document}

\maketitle

\vspace{-2em}
\begin{center}
  \twemoji{brain}\,
  \href{https://linhh29.github.io/blog/MAS-Gate/index.html}{Project page}
  \hspace{1em}
  \textcolor{red}{\faYoutube}\,
  \href{https://mas-gate-demo.hehailin.life/}{Online Demo}
  \hspace{1em}
  \faGithub\,
  \href{https://github.com/linhh29/MAS_Gate}{Code}
\end{center}

\input{sections/0_abstract}
\input{sections/1_introduction}
\input{sections/2_related_work}
\input{sections/3_quantification}
\input{sections/4_lifecycle_applications}
\input{sections/5_conclusion}
\input{sections/6_statements}

\bibliography{iclr2027_conference}
\bibliographystyle{iclr2027_conference}

\input{sections/A_appendix}

\end{document}

%% file: math_commands.tex
\usepackage{amsmath,amsfonts,bm}

\def\eqref#1{equation~\ref{#1}}

\def\1{\bm{1}}

\DeclareMathAlphabet{\mathsfit}{\encodingdefault}{\sfdefault}{m}{sl}
\SetMathAlphabet{\mathsfit}{bold}{\encodingdefault}{\sfdefault}{bx}{n}



%% file: sections/0_abstract.tex
\begin{abstract}
Large language model (LLM)-based multi-agent systems (MAS) can solve complex problems beyond the reach of a single agent by decomposing tasks and coordinating complementary capabilities. Despite extensive research on MAS, existing work largely focuses on specific aspects or individual phases of their development and operation, resulting in fragmented perspectives with objectives and signals that are difficult to reconcile across these different settings. To bridge these disconnected perspectives, we conceptualize MAS development and operation as a three-phase lifecycle spanning design-time orchestration \emph{before execution}, runtime intervention \emph{during execution}, and post-hoc attribution \emph{after execution}, and introduce MAS-Gate, a unified node-centric quantification applicable across these phases. Modeling each execution as a directed acyclic graph, MAS-Gate combines within-node utility, which estimates a node's contribution through changes in reference-answer likelihood, with inter-node strength, which approximates semantic uptake and potential downstream reach. We adapt this shared quantification to three objectives: providing a structure-aware dense reward for Orchestrator training, training an Editor to act on execution prefixes using paired rewards from completed edited and unedited trajectories, and supplying risk-ranked nodes as auxiliary evidence for LLM-based post-hoc attribution. Experiments spanning interaction environments, knowledge-intensive multi-hop question answering, general reasoning, and post-hoc attribution demonstrate that MAS-Gate improves workflow construction, targeted runtime editing, and LLM-assisted error localization over corresponding phase-specific baselines. Together, these results establish MAS-Gate as a unified structural interface for improving MAS generation, dynamic editing, and error attribution throughout the lifecycle.
\end{abstract}

%% file: sections/1_introduction.tex
\section{Introduction}
\label{sec:introduction}

Multi-agent systems (MAS) have emerged as an effective way to improve the performance of large language models (LLMs) on complex tasks~\citep{ke2025survey, huang2026ama, tran2025multi, lin2026interactive, jwalapuram2026illusion, yu2026beyond}. By decomposing a problem, assigning specialized roles, and coordinating intermediate results, an MAS can elicit complementary capabilities that are difficult to obtain from a single model invocation~\citep{yang2026understanding, lin2026unified, chen2024agentverse, cemri2026multi}. This potential has motivated research on many aspects of MAS, including workflow construction, topology search, model routing, communication pruning, and failure diagnosis in multi-agent executions~\citep{zhang2025aflow, zhuge2024language, yue2025masrouter, liu2023dynamic}. Despite this breadth, existing methods remain largely siloed, with most optimizing or analyzing a particular phase through a phase-specific objective, representation, or heuristic~\citep{hu2025automated, ke2025mas}. This specialization produces signals tailored to individual objectives, with limited correspondence across orchestration, intervention, and attribution~\citep{qi2026beyond}.

To organize these phase-specific objectives, we define the MAS lifecycle as an integrated framework spanning the initial construction of an MAS, its dynamic adjustment during execution, and the attribution of failures after execution~\citep{li2024survey, qi2026beyond}. Specifically, \emph{Before execution (design-time orchestration)}, the goal is to construct an effective MAS by organizing its agents and their interactions~\citep{fan2025workflowllm, ruan2026aorchestra}. \emph{During execution (runtime intervention)}, the goal is to monitor the partial trajectory and selectively adjust the active MAS, e.g., adding or removing sub-agents, replacing an agent, or rerouting dependencies without regenerating the entire workflow. \emph{After execution (post-hoc attribution)}, the goal is to analyze the completed wrong execution and identify the components responsible for its failure~\citep{xia2026broke, zhu2025llm}. Existing research has concentrated primarily on the first and third phases, whereas runtime intervention remains comparatively underexplored. Some systems respond to execution failures by regenerating or refining the entire MAS~\citep{wang2026mas}, but we focus on using the accumulated trajectory to selectively and dynamically adjust the continuation of an active multi-agent execution.

\begin{figure*}[t]
    \centering\includegraphics[width=\linewidth]{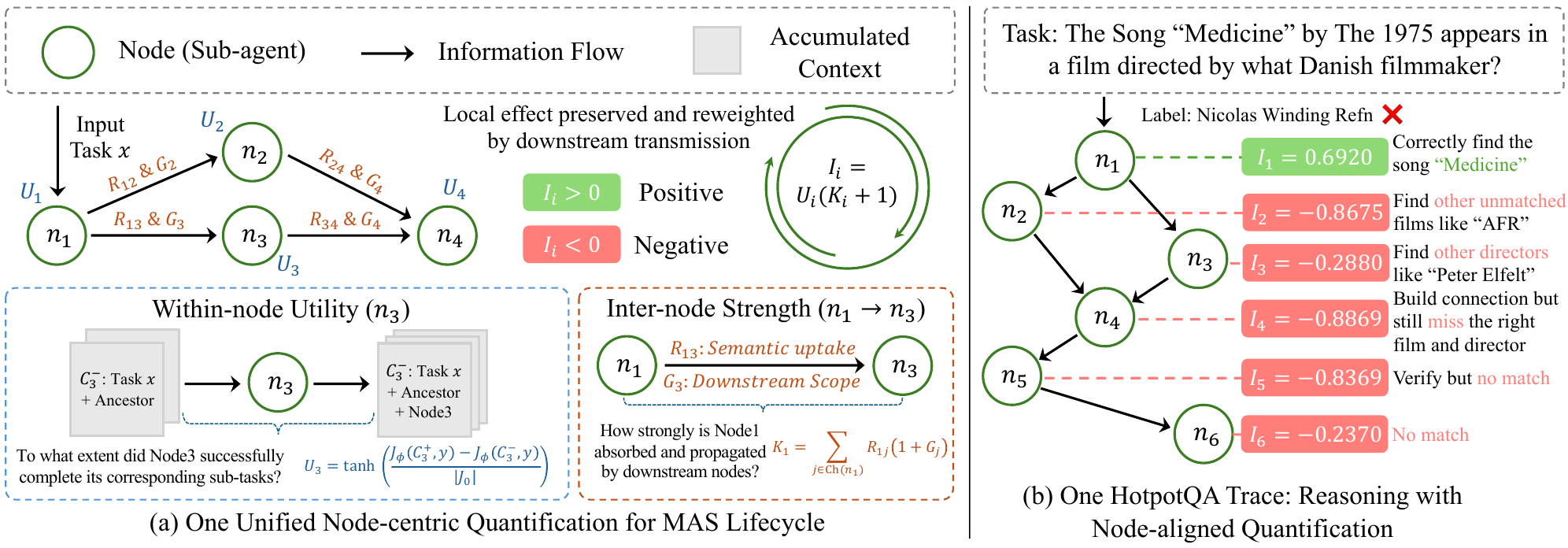}
    \caption{Overview of MAS-Gate. (a) Unified node-centric quantification combines within-node utility and inter-node strength to measure node impact across the MAS lifecycle. (b) A HotpotQA trace illustrates how node impact scores distinguish useful contributions from erroneous reasoning.}
    \label{fig:background}
\end{figure*}

In this paper, we introduce MAS-Gate, a unified node-centric quantification that provides a common formulation for the distinct objectives of the MAS lifecycle.
Following previous works~\citep{zhuge2024language}, we model the MAS as a directed acyclic graph to expose dependencies among agent computations. Each node represents a sub-agent, and each directed edge records which upstream outputs are available to a later sub-agent. MAS-Gate then quantifies two complementary properties of this execution through a \emph{Node-Centric Quantification}. First, within-node utility $U_i$ quantifies the change in reference-answer likelihood induced by incorporating node $i$'s output, serving as a proxy for its contribution to task progress. Second, inter-node strength $K_i$ combines estimated semantic uptake with potential downstream reach. The resulting node impact $I_i=U_i(1+K_i)$ integrates these complementary perspectives while preserving local utility when $K_i=0$. Figure~\ref{fig:background}(a) summarizes this quantification, while Figure~\ref{fig:background}(b) illustrates how the resulting node impact scores reflect the quality of a HotpotQA trace. In this example, the initial step correctly identifies the song and receives a positive impact score, whereas subsequent steps pursue incorrect film and director associations and fail to recover the correct answer, receiving negative scores. MAS-Gate thus distinguishes a useful intermediate contribution from erroneous reasoning within the same failed execution.

We evaluate the same MAS-Gate quantification across all three phases of the MAS lifecycle. For design-time orchestration, we combine mean node impact with terminal correctness as a dense reward to train an Orchestrator, improving the initial MAS workflow construction ability. For runtime intervention, we train an Editor to condition on the current MAS graph and accumulated execution trace, applying targeted adjustments to the remaining workflow after each node completes and before its output propagates downstream. For post-hoc error attribution, we derive risk scores from node utility and inter-node strength to rank the top-$k$ suspicious steps or agents as auxiliary evidence for an LLM localizer. The localizer retains access to the complete execution trajectory and is free to identify candidates beyond the supplied ranking.

Across experiments, Qwen3-8B~\citep{qwen3technicalreport} serves as the backbone for training the Orchestrator and Editor, Qwen3-30B-A3B-Instruct-2507 serves as the frozen sub-agent model for sub-task execution and within-node utility $U_i$ quantification, and bge-m3-zeroshot-v2.0~\citep{laurer_building_2023} provides the entailment scores used to quantify inter-node strength $K_i$. The evaluation covers interaction environment, multi-hop question answering, general reasoning, and post-hoc error attribution, with the first three task families supporting the design-time and runtime studies and the last supporting the attribution experiment. MAS-Gate outperforms existing orchestration baselines in task performance and generalization, and improves failure localization over direct LLM-based attribution. Beyond these established settings, our runtime intervention study demonstrates that targeted adjustments to an active MAS can improve execution outcomes without reconstructing the entire workflow, supporting dynamic intervention as an effective means of refining multi-agent execution. Together, these findings support the use of a common quantification mechanism to guide distinct objectives across the MAS lifecycle. Our contributions are as follows:
\begin{itemize}[leftmargin=*]
    \item We formulate the MAS lifecycle as a unified framework encompassing design-time orchestration, runtime intervention, and post-hoc attribution, connecting different existing MAS research.
    \item We introduce MAS-Gate, a unified node-centric quantification method that combines within-node utility and inter-node strength, providing a shared quantitative signal that can be adapted across all three lifecycle phases.
    \item We evaluate MAS-Gate across interactive environments, multi-hop question answering, general reasoning, and failure attribution, demonstrating gains over existing baselines in orchestration and failure localization, as well as the effectiveness of targeted intervention during execution.
\end{itemize}

%% file: sections/2_related_work.tex
\section{Related Work}
\label{sec:related_work}

\subsection{The Multi-Agent System Lifecycle}

\textbf{Before Execution: Design-Time Orchestration.}
Design-time orchestration constructs the agents, roles, communication structure, and execution order that will carry out a task~\citep{guo2024large, wu2026memory, qi2026beyond}. Existing approaches can be broadly divided into inference-time orchestration and training-time orchestration~\citep{ke2025survey}. Inference-time methods generate or search for a workflow without updating the underlying agent models. They use language-model-guided workflow synthesis, heuristic or evolutionary search, role and model assignment, and graph or topology construction to determine the decomposition and dependencies of an MAS~\citep{zhang2025aflow, ruan2026aorchestra, ke2025mas, li2026ofa, zhang2024g}. Training-time orchestration instead trains a dedicated orchestrator or workflow policy to emit, score, or refine MAS specifications from task feedback and execution traces. Its representative routes include orchestration-trace policy optimization, orchestration-level reward modeling, counterfactual credit assignment for workflow decisions, and trainable architecture or topology search~\citep{zhang2026reinforcement, chen2026lemon, tsang2026reward, ke2026mas, zhang2025multi, shi2026learning}. These approaches seek to learn or construct an orchestrator that produces more effective multi-agent workflows.

\textbf{During Execution: Runtime Intervention.}
Runtime intervention adjusts an ongoing execution using intermediate observations. Self-Refine improves an existing output through iterative feedback and revision by the same LLM~\citep{madaan2023selfrefine}. AgentDropoutV2 intercepts agent outputs and uses retrieved failure indicators to rectify erroneous messages or reject those that remain unreliable~\citep{wang2026agentdropoutv2}. Related approaches optimize agent communication: DyLAN combines agent-team selection with dynamic collaboration~\citep{liu2023dynamic}, while AgentPrune performs one-shot pruning of spatial and temporal communication dependencies~\citep{zhang2025agentprune}.
Building on these directions, We study runtime intervention through an Editor that conditions on the current MAS graph and accumulated execution trace to selectively revise the remaining workflow while preserving unaffected computation. The central challenge is to determine whether and where intervention is needed as weaknesses emerge during execution, enabling targeted adjustments before problematic outputs propagate downstream.

\textbf{After Execution: Post-Hoc Attribution.}
Post-hoc attribution focuses on failed MAS trajectories and seeks to identify the agent, node, or dependency responsible for the observed error. The Who\&When benchmark is designed specifically for this setting, pairing multi-agent execution traces with the question of which agent introduced or propagated the failure~\citep{zhang2025agent}. AEGIS complements this setting by injecting controlled errors into successful multi-agent executions and annotating the faulty agents and their error modes~\citep{kong2026aegis}. Related methods use spectrum-based analysis, trace-level failure localization, failure-mode taxonomies, graph-based execution analysis, and counterfactual contribution estimates to rank suspicious agents or interactions~\citep{xia2026broke, kong2026aegis, ge2026spectrum, ma2026safety}. The central output of this phase is an attribution of responsibility within an already completed and erroneous execution.

\textbf{Lifecycle Perspective.}
Existing methods are typically developed and evaluated within individual lifecycle phases, yielding representations and signals tailored to their respective objectives. These outputs are therefore not directly interchangeable across orchestration, intervention, and attribution. MAS-Gate connects these settings through a common node-scoring formulation.

\subsection{Quantification of Multi-Agent Systems}

Prior work quantifies multi-agent interaction with information-theoretic measures of uncertainty and information exchange, as well as graph-based measures of diversity, redundancy, communication efficiency, and structural connectivity~\citep{kim2025towards, yang2026understanding, lee2025gemmas, shen2025understanding,chen2026every, rath2026agent}. Quantification methods based on repeated rollouts or counterfactual interventions require additional MAS executions to estimate agent contributions, introducing computational overhead that can become prohibitive during large-scale training or latency-sensitive deployment. In contrast, MAS-Gate provides node-centric quantification through a single forward pass over the observed execution trace, without re-executing the MAS.

%% file: sections/3_quantification.tex
\section{Node-Centric Quantification}
\label{sec:quantification}

\subsection{Execution Representation}
\label{sec:execution_representation}

Given a task $x$ and reference answer $y$, we represent an MAS execution as a directed acyclic graph $\mathcal{G}=(V,E)$. Each node $i\in V$ corresponds to an accountable sub-agent that produces output $o_i$, and each edge $(i,j)\in E$ indicates that $o_i$ is available to node $j$. The resulting execution trace is
\begin{equation}
    \tau = \bigl(x,y,\mathcal{G},\{o_i\}_{i\in V},o_{\mathrm{final}}\bigr),
\end{equation}
where $o_{\mathrm{final}}$ denotes the system's final response. This representation captures both node-level outputs and the dependencies through which they may influence downstream computation.
Building on this representation, MAS-Gate characterizes each execution along two complementary dimensions. Within-node utility estimates local contribution through changes in reference-answer support, whereas inter-node strength captures semantic uptake and potential downstream reach. Their combination yields a node-level impact score that relates local utility to the semantic and structural context of subsequent computation.

\subsection{Within-Node Utility Quantification}
\label{sec:within_node_utility}

Dynamically generated subtasks rarely admit gold-standard intermediate labels, making it difficult to assign reliable dense rewards to individual sub-agents. Terminal correctness provides only sparse execution-level supervision and does not reveal whether a particular node moves the accumulated trajectory toward or away from the reference answer. We therefore adopt a perplexity-guided reward as a proxy to measure this effect~\citep{lin2026unified, liu2026rectifying}. The underlying premise is that a useful node output should make the reference answer $y$ easier to predict from the evidence accumulated so far, whereas a harmful output should have the opposite effect.

Let $C$ denote the accumulated evidence available to a sub-agent, and let $p_{\phi}$ denote the sub-agent LLM. We use this LLM to evaluate the predictability of the reference answer $y$ given $C$ and define
\begin{align}
    J_{\phi}(C,y)
    &= \frac{1}{|y|}\sum_{t=1}^{|y|}
    \log p_{\phi}\!\left(y_t\mid C,y_{<t}\right),\\
    \operatorname{PPL}_{\phi}(y\mid C)
    &= \exp\!\left(-J_{\phi}(C,y)\right),
\end{align}
where $J_{\phi}(C,y)$ is the mean token-level log-likelihood assigned by the sub-agent LLM to $y$ conditioned on $C$. A larger $J_{\phi}(C,y)$ corresponds to lower perplexity and indicates that the accumulated evidence better supports the reference answer. Let $\operatorname{Anc}(i)$ denote the set of ancestors of node $i$ in $\mathcal{G}$. For node $i$, the pre-node context contains the task and the outputs of these ancestors, while the post-node context additionally includes $o_i$:
\begin{align}
    C_i^{-} &= \text{Concat}\!\left(x,\{o_a:a\in\operatorname{Anc}(i)\}\right),\\
    C_i^{+} &= \text{Concat}\!\left(x,\{o_a:a\in\operatorname{Anc}(i)\},o_i\right).
\end{align}
To normalize the change against a task-only reference, we define $J_0=J_{\phi}(x,y)$ to measure the sub-agent LLM's baseline log-likelihood for the reference answer before any agent output is provided. The within-node utility is then defined as
\begin{equation}
    U_i = \tanh\!\left(
        \frac{
            J_{\phi}(C_i^{+},y)-J_{\phi}(C_i^{-},y)
        }{
            \left|J_0\right|
        }
    \right).
    \label{eq:u_i}
\end{equation}

If adding $o_i$ increases the log-likelihood of $y$, it lowers the corresponding perplexity and produces a larger $U_i$, and vice versa. Normalization by the task-only baseline improves comparability across tasks, while the $tanh$ function smooths outliers and preserves the ordering of node contributions.

\subsection{Inter-Node Strength Quantification}
\label{sec:information_uptake}

Within-node utility characterizes local reference-answer support but does not account for the node's downstream context. Inter-node strength incorporates this context through two factors: estimated semantic uptake by each child $j$, denoted by $R_{ij}$, and the downstream scope, denoted by $G_j$.

For each edge $(i,j)\in E$, we use entailment between the upstream output $o_i$ and downstream output $o_j$ as a semantic proxy for uptake:
\begin{equation}
    R_{ij}=p_{\mathrm{NLI}}\!\left(\mathrm{entailment}\mid o_i,o_j\right),
    \qquad R_{ij}\in[0,1].
    \label{eq:r_ij}
\end{equation}
where $p_{\mathrm{NLI}}$ is an entailment model and $R_{ij}$ is the entailment probability for the pair $(o_i,o_j)$. 
We also consider its potential downstream reach. Let $\operatorname{Ch}(i)$ and $\operatorname{Desc}(i)$ denote the sets of children and descendants of node $i$, respectively. We define the propagation scope of node $j$ as
\begin{equation}
    G_j =
    \frac{|\operatorname{Desc}(j)|}
    {\max_{v\in V}|\operatorname{Desc}(v)|}.
    \label{eq:g_j}
\end{equation}
Since each execution graph contains at least one dependency edge, the denominator is strictly positive. $G_j=0$ for a terminal node indicates no further reach, rather than no semantic uptake at that node. We therefore apply a unit baseline shift to $G_j$, so that an edge ending at a terminal child still contributes $R_{ij}$ to the strength of its predecessor. The resulting inter-node strength is
\begin{equation}
    K_i = \sum_{j\in\operatorname{Ch}(i)}R_{ij}(1+G_j).
    \label{eq:k_i}
\end{equation}
We then combine local utility and inter-node strength to define node impact. A second unit baseline shift is introduced at this stage to preserve the node's local contribution independently of downstream transmission: $I_i = U_i(1+K_i).$
A node with large $U_i$ and $K_i$ provides useful information that is broadly incorporated into subsequent computation. In contrast, a node with low or negative $U_i$ but large $K_i$ indicates weak or harmful information that is amplified downstream.


%% file: sections/4_lifecycle_applications.tex
\section{Lifecycle Applications of Node-Centric Quantification}
\label{sec:lifecycle_applications}


\subsection{Design-Time Orchestration}
\label{sec:design_time_orchestration}

\subsubsection{Experimental Settings}

\textbf{Dense reward for Orchestrator training.}
For an execution trace $\tau$ with realized node set $V$, we average the node impacts to obtain an MAS-level dense reward and combine it with binary answer correctness:
\begin{equation}
    r(\tau)=\frac{1}{|V|}\sum_{i\in V}I_i+c(\tau),
    \label{eq:design_reward}
\end{equation}
where $c(\tau)\in\{0,1\}$ denotes the terminal accuracy reward. The resulting objective retains outcome-level supervision while providing a graded signal that reflects the contributions of individual nodes throughout the execution. We use this fused reward with standard group-relative policy optimization (GRPO)~\citep{shao2024deepseekmath} to train the Orchestrator, while leaving the sub-agents and quantification models fixed. Appendix~\ref{app:ablation_reward} compares this fused reward with an accuracy-only reward, demonstrating the contribution of MAS-Gate quantification to orchestration performance and out-of-task generalization.

\textbf{Benchmarks.}
We consider three task families: interaction environment, multi-hop question answering, and general reasoning. The Orchestrator is trained on ALFWorld~\citep{shridhar2020alfworld}, HotpotQA~\citep{yang2018hotpotqa}, and DeepScaler~\citep{tan2025deepscaler}. We evaluate its in-task performance on ALFWorld, HotpotQA, and AIME~\citep{aime}, and its out-of-task (OOT) generalization on SciWorld~\citep{wang2022scienceworld}, BrowseComp-Plus~\citep{chen2025browsecomp}, and GPQA-Diamond~\citep{rein2023gpqa}. Here, OOT evaluation assesses transfer to benchmarks not used for Orchestrator training. All metrics are reported as 0-100\% percentages. Appendix~\ref{app:design_experiments} provides more details on the datasets, data splits, and evaluation protocols.

\textbf{Baselines.}
We compare MAS-Gate against three baseline families. (1)~CoT~\citep{wei2022chain} and CoT-SC~\citep{wang2022self} serve as general inference baselines; (2)~AFlow~\citep{zhang2025aflow}, AOrchestra~\citep{ruan2026aorchestra}, and Skill-MAS~\citep{lin2026skill} represent inference-time workflow construction; and (3)~MAS-GPT~\citep{ye2025mas}, ToolOrchestra~\citep{su2025toolorchestra}, and MAS-Orchestra~\citep{ke2026mas} represent training-time approaches that learn multi-agent coordination. Detailed descriptions and adaptation settings for all baselines are illustrated in Appendix~\ref{app:baselines}.

\begin{table*}[t]
\begin{center}
\caption{Quantification comparison of MAS-Gate against baselines on design-time orchestration. ALFW, SciW, VITA, BCP, and GPQA denote ALFWorld, SciWorld, VitaBench, BrowseComp-Plus, and GPQA-Diamond, respectively. \textbf{Bold} indicates the best result.}
\label{tab:main_results}
\setlength{\tabcolsep}{2.8pt}
\resizebox{\textwidth}{!}{%
\begin{tabular}{llcccccccc}
\toprule
\multirow{2.5}{*}{Category} & \multirow{2.5}{*}{Method} & \multicolumn{3}{c}{Train: ALFW} & \multicolumn{2}{c}{Train: HotpotQA} & \multicolumn{2}{c}{Train: DeepScaler} & \multirow{2.5}{*}{Avg} \\
\cmidrule(lr){3-5} \cmidrule(lr){6-7} \cmidrule(lr){8-9} 
 &  & ALFW & SciW\textsubscript{OOT} & VITA\textsubscript{OOT} & HotpotQA & BCP\textsubscript{OOT} & AIME & GPQA\textsubscript{OOT} & \\
\midrule

\multirow{2}{*}{General} & CoT & 37.18 & 1.26 & 29.19 & 50.00 & 11.31 & 28.07 & 48.99 & 29.43 \\
 & CoT-SC & 32.05 & 1.07 & 28.23 & 54.00 & 8.33 & 29.82 & 52.02 & 29.36 \\
\midrule
\multirow{3}{*}{Inference-time} & AFlow & 31.09 & 3.90 & 23.54 & 46.50 & 12.50 & 15.79 & 33.84 & 23.88 \\
 & AOrchestra & 57.05 & 8.93 & 9.18 & 48.00 & 12.50 & 40.35 & 57.07 & 33.30 \\
 & Skill-MAS & 28.21 & 5.25 & 13.55 & 54.00 & 8.93 & 19.30 & 48.99 & 25.46 \\
\midrule
\multirow{3}{*}{Training-time} & MAS-GPT & 53.53 & 14.81 & 7.52 & 51.00 & 11.31 & 24.56 & 54.04 & 30.97 \\
 & ToolOrchestra & 37.50 & 7.69 & 10.98 & 37.50 & 3.57 & 12.28 & 36.36 & 20.84 \\
 & MAS-Orchestra & 14.42 & 2.33 & 22.22 & 51.00 & 9.52 & 21.05 & 41.41 & 23.14 \\
\midrule
MAS-Gate & Orchestrator & \textbf{57.69} & \textbf{18.27} & \textbf{45.29} & \textbf{69.50} & \textbf{13.69} & \textbf{47.37} & \textbf{61.11} & \textbf{44.70} \\
\bottomrule
\end{tabular}%
}
\end{center}
\end{table*}

\subsubsection{Results and Analysis}

First, MAS-Gate delivers consistent performance gains over both single-agent and multi-agent baselines. As shown in Table~\ref{tab:main_results}, existing MAS baselines do not consistently outperform single-agent baselines. For example, AOrchestra surpasses CoT on ALFWorld and ScienceWorld but falls below it on VitaBench. Such variation is consistent with prior findings that architectural inefficiencies can prevent automatically constructed MAS from realizing the benefits of multi-agent coordination~\citep{jwalapuram2026illusion}. In contrast, MAS-Gate achieves the best results across all seven evaluation benchmarks, reaching an average score of 44.70 against 33.30 for the strongest competing MAS method. These gains support the value of MAS-Gate quantification that guides the Orchestrator toward more effective multi-agent organization.

Second, MAS-Gate's orchestration policy generalizes beyond its training benchmarks. It achieves the best performance on all four OOT benchmarks, with the largest margin over competing MAS methods on VitaBench. On this benchmark, all six MAS baselines fall below CoT, whereas MAS-Gate reaches 45.29\%, surpassing the strongest MAS baseline by 21.75\%. This contrast highlights the limited transferability of existing MAS methods, whose advantages do not reliably carry over to unseen benchmarks. With MAS-Gate quantification as a structure-aware training reward, the Orchestrator instead maintains a consistent advantage across OOT settings. Taken together, these results support the transferability of MAS-Gate-trained orchestration across the evaluated task settings, without identifying the specific coordination mechanisms underlying this advantage.

\subsection{Runtime Intervention}
\label{sec:runtime_intervention}

\subsubsection{Experimental Settings}

\textbf{Paired reward for Editor training.}
We use MAS-Gate to train an Editor for selective runtime intervention, keeping the Orchestrator and sub-agents fixed. After each node completes and before downstream nodes consume its output, the Editor observes the task, the current MAS graph, and the accumulated execution outputs. It can choose from six actions, grouped by their position relative to the current node: (1)~\emph{Current-node decisions}: leave execution unchanged (\textsc{Keep}), retry the current node, or append a bounded diagnostic annotation to the current node's output; (2)~\emph{Upstream intervention}: re-execute a parent and its affected descendants; and (3)~\emph{Downstream intervention}: guide an unexecuted child or insert an auxiliary sub-task into the remaining workflow. All actions are restricted to legal targets and preserve unaffected computation. To balance computational efficiency and execution stability, we allow at most one non-\textsc{Keep} intervention over the entire execution trace.


During training, we compare an edited execution $\tau^a$ with a cached no-intervention reference $\tau^0$ for the same task and initial workflow. Both executions share the observed prefix up to the intervention decision. The Editor selects actions from the observed prefix, whereas the paired reward is evaluated after both continuations are complete. The decision interface is specified in terms of the task, graph, and observed outputs; reference answers and quantification scores are reserved for reward computation (Appendix~\ref{app:implementation_details_exp2}). We aggregate the shared node impacts as
$\bar{I}(\tau)=|V_\tau|^{-1}\sum_{i\in V_\tau} I_i$
where $V_\tau$ is the realized node set. A valid non-\textsc{Keep} action $a$ receives the paired reward
\begin{equation}
\begin{aligned}
    r(a) ={}
    c(\tau^a)-c(\tau^0)
    + \beta\bigl[\bar{I}(\tau^a)-\bar{I}(\tau^0)\bigr]
    - \lambda C(a).
\end{aligned}
\label{eq:runtime_reward}
\end{equation}
where $c(\tau)$ is the task outcome score, $C(a)$ is a predefined action-dependent intervention penalty, and $\beta$ and $\lambda$ control the quantification and cost terms. This reward balances task improvement, changes in node-level contributions, and intervention cost. Each \textsc{Keep} decision receives zero local reward, while invalid actions receive a fixed negative penalty.
We train the Editor using GRPO~\citep{shao2024deepseekmath} with multiple decision sequences sampled for each task. 

\textbf{Benchmarks.}
We adopt the training and evaluation benchmarks from Section~\ref{sec:design_time_orchestration}, excluding VitaBench. It involves multi-turn interactions, and each turn requires multiple tool calls, which does not readily support the paired comparisons required here, where the no-intervention and Editor branches must share the same initial workflow and pre-intervention execution prefix. Further details on the benchmarks and data splits are provided in Appendix~\ref{app:design_experiments}.

\textbf{Baselines.}
We compare against three baseline families. (1)~\emph{LLM Editors} use DeepSeek-V4-Flash and GPT-5.4-Mini as frozen prompted Editors to intervene in the underlying MAS. (2)~\emph{Output Correction} includes Self-Refine~\citep{madaan2023selfrefine}, which iteratively improves outputs through self-feedback, and AgentDropoutV2~\citep{wang2026agentdropoutv2}, which rectifies erroneous agent outputs or rejects those that remain unreliable. (3)~\emph{Structural Adaptation} includes DyLAN~\citep{liu2023dynamic}, which combines agent-team selection with dynamic collaboration, and AgentPrune~\citep{zhang2025agentprune}, which prunes spatial and temporal communication dependencies. Detailed baseline descriptions are provided in Appendix~\ref{app:runtime_baselines} and implementation details in Appendix~\ref{app:implementation_details}.

\begin{table*}[t]
\begin{center}
\caption{Performance comparison of MAS-Gate against baselines on runtime intervention. ALFW, SciW, BCP, and GPQA denote ALFWorld, SciWorld, BrowseComp-Plus, and GPQA-Diamond, respectively. \textbf{Bold} indicates the best result.}
\label{tab:runtime_results}
\setlength{\tabcolsep}{2.8pt}
\resizebox{\textwidth}{!}{%
\begin{tabular}{llccccccc}
\toprule
\multirow{2.5}{*}{Category}
& \multirow{2.5}{*}{Method}
& \multicolumn{2}{c}{Train: ALFW}
& \multicolumn{2}{c}{Train: HotpotQA}
& \multicolumn{2}{c}{Train: DeepScaler}
& \multirow{2.5}{*}{Avg} \\
\cmidrule(lr){3-4} \cmidrule(lr){5-6} \cmidrule(lr){7-8}
& & ALFW & SciW\textsubscript{OOT}
& HotpotQA & BCP\textsubscript{OOT}
& AIME & GPQA\textsubscript{OOT} & \\
\midrule
\multirow{2}{*}{\shortstack[l]{LLM\\Editors}}
& DeepSeek-V4-Flash & 34.94 & 7.33 & 62.00 & 11.31 & 31.58 & 54.04 & 33.53 \\
& GPT-5.4-Mini & 37.50 & 5.67 & 57.50 & 9.52 & 35.09 & 59.09 & 34.06 \\
\midrule
\multirow{2}{*}{\shortstack[l]{Output\\Correction}}
& Self-Refine & 48.40 & 13.67 & 58.00 & 5.36 & 38.60 & 55.05 & 36.51 \\
& AgentDropoutV2 & 53.21 & 17.00 & 69.00 & 9.52 & 42.11 & 63.64 & 42.41 \\
\midrule
\multirow{2}{*}{\shortstack[l]{Structural\\Adaptation}}
& DyLAN & 55.77 & 16.67 & 72.50 & 12.50 & 43.86 & 63.64 & 44.16 \\
& AgentPrune & 51.28 & 9.34 & 70.00 & 10.12 & 31.58 & 61.11 & 38.91 \\
\midrule
MAS-Gate & Editor
& \textbf{59.62} & \textbf{22.33}
& \textbf{75.50} & \textbf{15.48}
& \textbf{49.12} & \textbf{65.15}
& \textbf{47.87} \\
\bottomrule
\end{tabular}%
}
\end{center}
\end{table*}

\subsubsection{Results and Analysis}

Table~\ref{tab:runtime_results} shows that MAS-Gate consistently outperforms prompted LLM Editors, output-correction methods, and structural-adaptation methods across all six benchmarks, achieving an average score of 47.87 compared with 44.16 for the strongest baseline. Specifically, under the same Editor protocol, the trained Qwen3-8B Editor substantially outperforms the frozen prompted DeepSeek-V4-Flash and GPT-5.4-Mini Editors. This comparison suggests that effective intervention in an ongoing MAS requires more than general reasoning ability and benefits from execution-level feedback that teaches the Editor when and how to adjust the workflow. Furthermore, the proposed online intervention setting outperforms design-time orchestration alone (Table~\ref{tab:main_results}) on all six shared benchmarks, highlighting the value of extending MAS optimization into execution, where emerging evidence enables targeted adjustments unavailable at design time. Together, these results support MAS-Gate quantification as a useful supervisory signal for learning this intervention capability, enabling a compact Editor to make effective runtime decisions without access to quantification scores at inference time.

\subsection{Post-Hoc Attribution}
\label{sec:post_hoc_attribution}

\subsubsection{Experimental Settings}

\textbf{Risk Score for Error Attribution.} For post-hoc error attribution, we represent each recorded trajectory as a chronological chain connecting consecutive steps. We use the negative part of the shared node impact as a risk score:
\begin{equation}
    s_i^{\mathrm{risk}}
    = \max(0,-I_i)
    = \max(0,-U_i)(1+K_i).
    \label{eq:attribution_risk}
\end{equation}
Negating $I_i$ converts adverse impacts into positive risk scores, so that steps with more negative impacts receive higher priority for attribution. Clipping at zero assigns zero risk to non-negative impacts, retaining only the component that indicates reduced support for the reference answer. We rank steps by this score for step-level attribution. For agent-level attribution, we use the maximum risk over an agent's eligible steps, so that a high-risk contribution is not averaged away by its other steps. The resulting top-$k$ ranking is provided to an LLM localizer as auxiliary evidence.

\textbf{Benchmarks.}
We evaluate two attribution tasks, i.e., step-level localization on Who\&When~\citep{zhang2025agent} and agent-level localization on the single-target test subset of AEGIS~\citep{kong2026aegis}.
Who\&When provides annotated error steps in failed executions, whereas
AEGIS identifies agents subjected to controlled error injection. We report exact step-level and agent-level accuracy for each benchmark. Further benchmark details are provided in Appendix~\ref{app:design_experiments}.

\textbf{Baselines.} 
We compare Direct LLM localization with its MAS-Gate-guided counterpart using the same complete execution trajectory and reference task answer. MAS-Gate additionally supplies a risk-based ranking of candidate steps or agents as auxiliary evidence. The comparison covers five localizers from the Qwen~\citep{qwen3technicalreport}, Mistral~\citep{jiang2023mistral7b}, and Llama~\citep{dubey2024llama} families. Appendix~\ref{app:error_attribution_baseline} and~\ref{app:implementation_details_exp3} provide the baseline definitions and implementation details.

\subsubsection{Results and Analysis}
Table~\ref{tab:failure_attribution} shows that MAS-Gate improves localization for four of five models on Who\&When and all five on AEGIS, supporting its applicability across model families and attribution granularities. Notably, the largest gain on each benchmark occurs for the localizer with the lowest Direct LLM accuracy: Llama-3.1-8B gains 5.43\% on Who\&When, and Qwen3.5-4B gains 5.84\% on AEGIS. This pattern suggests that risk-based guidance can help localizers who struggle to identify decisive evidence within a complete trajectory. Because both settings retain the same trajectory and reference answer, the gains highlight the value of explicitly prioritizing suspicious contributions within the available evidence. While Qwen3.6-27B shows a slight decline on Who\&When, MAS-Gate improves accuracy in all other model--benchmark pairs. This broad pattern supports the effectiveness of MAS-Gate quantification in guiding failure attribution across different LLMs and evaluation settings. The ablations in Appendix~\ref{app:error_attribution_ablation} further support this finding: risk-ranked candidates outperform both randomly selected candidates and the same candidates in shuffled order, indicating that both candidate selection and prioritization contribute to effective attribution.

\begin{table*}[t]
\caption{Performance comparison of MAS-Gate against Direct LLM on post-hoc error attribution. We report step-level accuracy on Who\&When and agent-level accuracy on the single-target AEGIS subset (\%). $\Delta$ denotes the change over Direct LLM in percentage points. \textbf{Bold} indicates the better result across pairs.}
\label{tab:failure_attribution}
\begin{center}
\setlength{\tabcolsep}{2.8pt}
\resizebox{\textwidth}{!}{%
\begin{tabular}{lcccccc}
\toprule
\multirow{2.5}{*}{Localizer}
& \multicolumn{3}{c}{Who\&When ($k=8$)}
& \multicolumn{3}{c}{AEGIS ($k=4$)} \\
\cmidrule(lr){2-4} \cmidrule(lr){5-7}
& Direct LLM & + MAS-Gate & $\Delta$
& Direct LLM & + MAS-Gate & $\Delta$ \\
\midrule
Qwen3-30B-A3B
& 33.70 & \textbf{37.50} & $+3.80$
& 43.07 & \textbf{43.80} & $+0.73$ \\
Qwen3.5-4B
& 34.24 & \textbf{35.87} & $+1.63$
& 34.31 & \textbf{40.15} & $+5.84$ \\
Qwen3.6-27B
& \textbf{40.76} & 39.67 & $-1.09$
& 52.55 & \textbf{54.01} & $+1.46$ \\
Mistral-Small-3.2-24B
& 30.98 & \textbf{32.07} & $+1.09$
& 50.36 & \textbf{51.82} & $+1.46$ \\
Llama-3.1-8B
& 21.20 & \textbf{26.63} & $+5.43$
& 38.69 & \textbf{40.15} & $+1.46$ \\
\bottomrule
\end{tabular}%
}
\end{center}
\end{table*}

\begin{wrapfigure}{r}{0.48\textwidth}
    \centering
    \includegraphics[width=\linewidth]{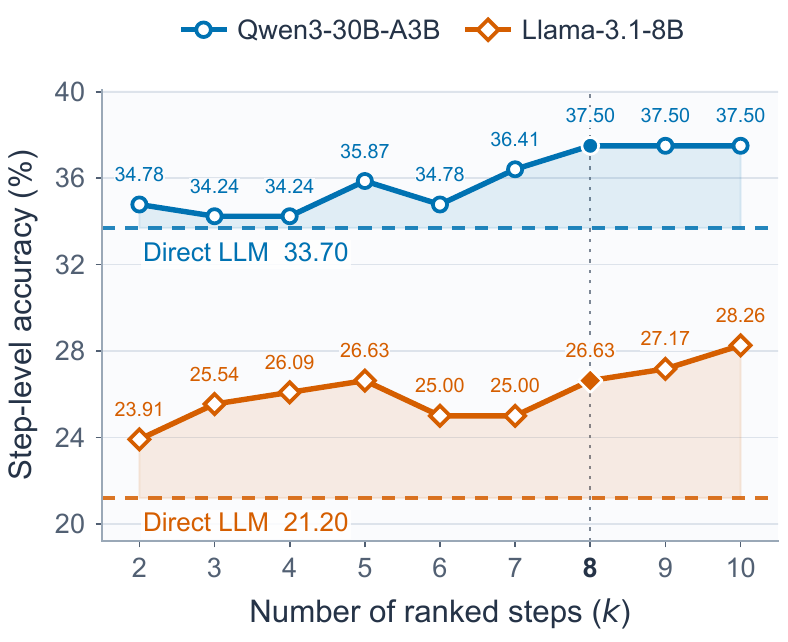}
    \caption{Effect of $k$ on Who\&When. Solid lines show MAS-Gate-guided localization and dashed lines show Direct LLM baselines.}
    \label{fig:who_when_topk}
\end{wrapfigure}
\noindent Figure~\ref{fig:who_when_topk} reveals two patterns in the effect of candidate count on Who\&When. First, MAS-Gate provides consistent benefits across candidate budgets: both localizers outperform Direct LLM for every tested $k$ from 2 to 10. The gains at $k=2$ further show that even a small set of prioritized steps can provide useful guidance within the complete trajectory. Second, the additional benefit of expanding this set depends on the localizer. Qwen3-30B-A3B reaches a plateau at $k=8$, whereas Llama-3.1-8B continues to improve from $k=8$ to $k=10$, despite fluctuations at smaller values. Thus, MAS-Gate remains effective across the tested range, but increasing the breadth of its guidance yields different returns for different models. Together, the improvements across localizers, the ranking ablations, and the sensitivity analysis support MAS-Gate quantification as an effective basis for prioritizing failure-relevant evidence and improving LLM-based post-hoc attribution.

%% file: sections/5_conclusion.tex
\section{Conclusion}
\label{sec:conclusion}

In this work, we introduce the MAS lifecycle as an execution-centered framework that unifies existing fragmented research on design-time orchestration, runtime intervention, and post-hoc attribution. We then propose MAS-Gate, a unified node-centric quantification that combines within-node utility and inter-node strength to provide structured feedback across all three phases. Experiments demonstrate improved orchestration and generalization, effective trained runtime intervention, and gains in assisted attribution across most evaluated settings, supporting the use of a shared quantification mechanism for these distinct objectives. The runtime results further highlight the benefits of adapting workflows to evidence that emerges during execution. Overall, MAS-Gate demonstrates the feasibility of a common quantification across the lifecycle. However, the three phases are currently optimized independently. Future work could explore how feedback from each phase can inform the others, leading to a more integrated MAS lifecycle. For example, the Orchestrator could learn from execution outcomes after Editor intervention, while the Editor adapts to recurring weaknesses in the generated workflows, allowing the two policies to improve together.

%% file: sections/6_statements.tex
\section*{Ethics Statement}

This work studies MAS orchestration, runtime intervention, and failure attribution in controlled research benchmarks. Our experiments use established datasets and model-generated execution traces, without recruiting human participants or collecting new personal or sensitive data. The evaluation concerns reasoning, question answering, and interactive tasks rather than decisions about individuals or demographic groups. We use all datasets in accordance with their intended research settings and licenses, and report our methodology and experimental results with an emphasis on transparency and reproducibility

\section*{Reproducibility Statement}

Our reinforcement-learning implementation uses \texttt{verl}. Appendix~\ref{app:pseudocode} presents the MAS-Gate quantification procedure, and Appendix~\ref{app:experimental_details} documents benchmarks, baselines, model configurations, training hyperparameters, and hardware. Appendix~\ref{app:prompt_details} provides prompts and output formats for the Orchestrator, Editor, and LLM-based evaluation components. We provide our implementation code in the supplementary material to facilitate reproduction of the experimental results across all three lifecycle phases.

\section*{LLM Usage}

LLM assistants were used during manuscript preparation to improve grammar, clarity, organization, and LaTeX formatting. All assisted text is manually reviewed by the authors, and the authors remain responsible for the final content.

%% file: sections/A_appendix.tex
\clearpage
\appendix

\section{Description of Appendix}
\label{app:overview}

This appendix provides methodological details and additional experimental evidence for MAS-Gate. Appendix~\ref{app:pseudocode} presents the shared node-centric quantification procedure and its use across the three lifecycle phases. Appendix~\ref{app:experimental_details} documents benchmark statistics, baseline configurations, and implementation details for each experiment. Appendix~\ref{app:additional_ablations} examines node-impact composition, the contribution of the design-time dense reward, and the roles of candidate selection and ranking in failure attribution. Appendix~\ref{app:case_study} illustrates workflow construction, runtime intervention, and post-hoc attribution through concrete execution traces. Finally, Appendix~\ref{app:prompt_details} provides the prompts and output formats used for orchestration, editing, failure localization, and benchmark evaluation.

\section{Pseudocode of MAS-Gate}
\label{app:pseudocode}

We provide the pseudocode of MAS-Gate here, which summarizes the common computation performed on an execution trace. 

\begin{algorithm}[htbp]
\caption{MAS-Gate node-centric quantification and lifecycle use}
\label{alg:mas_gate}
\begin{algorithmic}[1]
\Require Task $x$, reference $y$, execution DAG $\mathcal{G}=(V,E)$, outputs $\{o_i\}_{i\in V}$, scoring LLM $p_{\phi}$, NLI model $p_{\mathrm{NLI}}$, phase $s$
\Ensure Node records $\mathcal{S}$, edge scores $\{R_{ij}\}_{(i,j)\in E}$, and phase-specific signal
\State $J_0 \gets J_{\phi}(x,y)$
\For{$i$ in a topological ordering of $V$}
    \State $C_i^{-} \gets \text{Concat}\!\left(x,\{o_a:a\in\operatorname{Anc}(i)\}\right)$
    \State $C_i^{+} \gets \text{Concat}\!\left(x,\{o_a:a\in\operatorname{Anc}(i)\},o_i\right)$
    \State $U_i \gets \tanh\!\left((J_{\phi}(C_i^{+},y)-J_{\phi}(C_i^{-},y))/\left|J_0\right|\right)$
\EndFor
\For{$(i,j)\in E$}
    \State $R_{ij}\gets p_{\mathrm{NLI}}(\mathrm{entailment}\mid o_i,o_j)$
\EndFor
\State $D_{\max}\gets\max_{v\in V}|\operatorname{Desc}(v)|$
\Statex $D_{\max}>0$ since the execution graph contains at least one dependency edge.
\For{$j\in V$}
    \State $G_j\gets |\operatorname{Desc}(j)|/D_{\max}$
\EndFor
\For{$i\in V$}
    \State $K_i\gets\sum_{j\in\operatorname{Ch}(i)}R_{ij}(1+G_j)$
    \State $I_i\gets U_i(1+K_i)$
    \State Add $(i,U_i,K_i,I_i)$ and outgoing $R_{ij}$ values to $\mathcal{S}$
\EndFor
\If{$s=\textsc{Design}$}
    \State \Return $\mathcal{S}$ and $|V|^{-1}\sum_{i\in V}I_i+c(\tau)$
\ElsIf{$s=\textsc{Runtime}$}
    \State Cache reference $\tau^0$ and mean impact $\bar I(\tau^0)\gets |V|^{-1}\sum_i I_i$
    \State Sample $\pi_\theta$ on public execution prefixes; allow at most one non-\textsc{Keep} edit
    \State Set $r_t\gets 0$ for legal \textsc{Keep} decisions; $r_t\gets -0.25$ for invalid actions
    \If{a legal edit $a_t$ is selected}
        \State Apply $a_t$, complete and quantify $\tau^{a_t}$, then set $r_t\gets r(a_t)$ via Eq.~\ref{eq:runtime_reward}
    \EndIf
    \State \Return $\mathcal{S}$ and turn-local rewards $\{r_t\}$ for GRPO
\Else
    \State Compute step risks $s_i^{\mathrm{risk}}\gets\max(0,-I_i)$
    \State For agent-level attribution, aggregate $s_a^{\mathrm{risk}}\gets\max_{i\in\mathcal V_a^{\mathrm{elig}}}s_i^{\mathrm{risk}}$
    \State Rank eligible steps or agents by decreasing risk and retain the top-$k$ as $\mathcal R_k$
    \State \Return $\mathcal S$ and the predicted error step or agent from the localizer given $x,y$, full $\tau$, and $\mathcal R_k$
\EndIf
\end{algorithmic}
\end{algorithm}


\section{Experimental Details}
\label{app:experimental_details}

\subsection{Statistics of Benchmarks}
\label{app:design_experiments}

Table~\ref{tab:benchmark_statistics} summarizes how each benchmark is used across the experiments. ALFWorld, HotpotQA, and DeepScaleR serve as the three source benchmarks for training the design-time Orchestra on interactive tasks, multi-hop question answering, and general reasoning, respectively. Their associated in-task evaluations measure performance within the capability family represented in training. SciWorld, VitaBench, BrowseComp-Plus, and GPQA-Diamond are held-out out-of-task (OOT) benchmarks, which are used to measure the transfer performance to related but distinct task distributions.

\begin{table*}[t]
\caption{Benchmark roles and split statistics. OOT indicates that the benchmark is never used to update or select the Orchestra trained on the paired source task.}
\label{tab:benchmark_statistics}
\begin{center}
\small
\setlength{\tabcolsep}{4.2pt}
\begin{tabular}{p{0.13\textwidth}p{0.13\textwidth}p{0.15\textwidth}p{0.13\textwidth}p{0.13\textwidth}p{0.20\textwidth}}
\toprule
Task Family & Benchmark & Role & Train size & Test size & Task characteristic \\
\midrule
\multirow{6}{*}{\begin{tabular}[c]{@{}c@{}}Interactive\\ environments\end{tabular}} & ALFWorld & Train / In-task test & 4032 & 312 & Language-grounded household interaction \\
& SciWorld & OOT test only & - & 300 & Interactive scientific environments \\
& VitaBench & OOT test only & - & 100 & Multi-domain tool-use interaction \\
\midrule
\multirow{4}{*}{Multi-hop QA} & HotpotQA & Train / in-task test & 10000 & 200 & Multi-document evidence composition \\
& BrowseComp-Plus & OOT test only & - & 168 & Hard information aggregation \\
\midrule
\multirow{6}{*}{\begin{tabular}[c]{@{}c@{}}General\\ reasoning\end{tabular}} & DeepScaleR & Train only & 10000 & - & Complex mathematical reasoning \\
& AIME & In-task test only & - & 57 & Competition mathematics (Year 24\&25) \\
& GPQA-Diamond & OOT test only & - & 198 & Graduate-level science questions \\
\midrule
\multirow{4}{*}{\begin{tabular}[c]{@{}c@{}}Failure\\ attribution\end{tabular}}
& Who\&When & Evaluation only & - & 184
& Exact step-level attribution \\
& AEGIS & Evaluation only & - & 137
& Single-target agent-level attribution \\
\bottomrule
\end{tabular}
\end{center}
\end{table*}

\textbf{Interaction environments.} \textbf{ALFWorld} aligns abstract text environments with embodied household tasks~\citep{shridhar2020alfworld}. Its episodes require an agent to interpret a high-level language goal, plan over a long horizon, manipulate household objects, and adjust to the current layout and state. \textbf{SciWorld} places the agent in a text-based scientific environment where it must inspect objects, operate instruments, and conduct experiments through ordered actions before determining the scientific result~\citep{wang2022scienceworld}. \textbf{VitaBench} evaluates multi-turn tool-using agents in simulated service settings, including food delivery, in-store consumption, and online travel~\citep{he2026vitabench}. Its evolving user goals, dependent tool calls, and changing service states probe practical decision-making and recovery over extended interactions.

\textbf{Multi-hop QA.} \textbf{HotpotQA} is a Wikipedia-based multi-hop QA benchmark whose questions require evidence to be gathered and combined across multiple documents~\citep{yang2018hotpotqa}. \textbf{BrowseComp-Plus} evaluates deep-research agents on difficult information-seeking questions over a fixed, curated corpus~\citep{chen2025browsecomp}. The fixed corpus avoids dependence on the changing open web while preserving the need for iterative retrieval, source comparison, and evidence synthesis, making the benchmark appropriate for testing transfer from standard multi-hop QA to broader research-oriented reasoning.

\textbf{General reasoning.} The \textbf{DeepScaleR} training set contains competition-style mathematical problems collected for reinforcement-learning-based reasoning training~\citep{tan2025deepscaler}. The problems demand multi-step derivations but retain verifiable final answers, which provides a reliable outcome signal for optimization. \textbf{AIME} consists of challenging competition mathematics problems spanning algebra, geometry, combinatorics, and number theory~\citep{aime}. \textbf{GPQA-Diamond} comprises expert-written graduate-level questions in biology, physics, and chemistry~\citep{rein2023gpqa}. It tests scientific reasoning and specialized knowledge outside the mathematical distribution used for training, thereby serving as an OOT transfer evaluation.

\textbf{Failure attribution.}
For \textbf{Who\&When,} 
we use all 184 publicly released failure trajectories, comprising 126 algorithm-generated and 58 hand-crafted examples~\citep{zhang2025agent}. Each example contains the task query, reference answer, complete execution history, and an annotated error agent and step. We retain the released zero-based step indices and evaluate whether the predicted step exactly matches the annotated error step. Algorithm-generated trajectories contain 5-10 steps, with a mean of 8.72, while hand-crafted trajectories contain 5-130 steps, with a mean of 51.60.  \textbf{AEGIS} constructs failure trajectories through controlled error injection into multi-agent executions~\citep{kong2026aegis}. We use the test split and retain examples whose injection annotations identify exactly one distinct agent, yielding 137 of the 600 released test examples. The subset covers GAIA, GSM8K, SciBench, HumanEval, and MMLU-Pro, with trajectory lengths ranging from 3 to 43 steps and a mean of 13.83. The selection criterion concerns the number of distinct agents, rather than the number of injection annotations. We evaluate exact agent-level accuracy against the unique target agent. Model inputs retain the task query, reference answer, observed final output, and complete conversation history, but exclude injection indicators and error annotations.



\subsection{Baselines}
\label{app:baselines}

\subsubsection{Design-time orchestration}

\textbf{General inference methods.} \textbf{CoT} prompts the frozen LLM to produce intermediate reasoning before the final answer in one-shot~\citep{wei2022chain}. \textbf{CoT-SC} samples multiple reasoning trajectories and selects the answer with the highest consistency~\citep{wang2022self}. These methods do not construct an explicit multi-agent workflow and provide single-agent references for judging the value of learned coordination.

\textbf{Inference-time MAS methods.} \textbf{AFlow} searches over code-represented agentic programs and improves candidate workflows using Monte Carlo Tree search~\citep{zhang2025aflow}. \textbf{AOrchestra} instantiates task-specific sub-agents by configuring their instructions, context, tools, and models at inference time~\citep{ruan2026aorchestra}. \textbf{Skill-MAS} stores orchestration experience in an evolvable meta-skill and uses trajectory sampling and reflection to construct workflows without updating model parameters~\citep{lin2026skill}. These methods adapt the workflow at inference time rather than training an orchestrator.

\textbf{Training-time MAS methods.} \textbf{MAS-GPT} learns to generate executable, query-adaptive multi-agent programs~\citep{ye2025mas}. \textbf{ToolOrchestra} trains a lightweight orchestrator to coordinate models and tools under outcome-, efficiency-, and user-preference-aware rewards~\citep{su2025toolorchestra}. \textbf{MAS-Orchestra} formulates holistic MAS construction as function-calling reinforcement learning in which the orchestrator generates the complete organization~\citep{ke2026mas}. These are the closest design-time comparisons. These methods use distinct training objectives and orchestration formulations.



\subsubsection{Runtime intervention}
\label{app:runtime_baselines}

\textbf{LLM Editors.}
\textbf{DeepSeek-V4-Flash} and \textbf{GPT-5.4-Mini} serve as prompted editing controllers, while the underlying main-agent and sub-agent components are retained. Their role is to select runtime interventions from the observed execution state rather than directly replace the task-solving agents. This group compares general-purpose LLM prompting with an Editor specifically trained for runtime intervention using MAS-Gate feedback.

\textbf{Output Correction.}
\textbf{Self-Refine} iteratively improves a candidate output by having the same LLM generate feedback and revise its answer accordingly, without additional parameter training~\citep{madaan2023selfrefine, he2025self}. \textbf{AgentDropoutV2} intercepts intermediate agent outputs and retrieves failure indicators to guide error correction; outputs that remain unreliable are rejected to limit downstream error propagation~\citep{wang2026agentdropoutv2}. These methods focus on correcting or filtering generated content, providing output-level comparisons to our Editor, which can additionally select re-execution and workflow interventions.

\textbf{Structural Adaptation.}
\textbf{DyLAN} selects a task-oriented agent team based on agent contributions and supports dynamic collaboration during task solving~\citep{liu2023dynamic}. \textbf{AgentPrune} optimizes communication topology through one-shot pruning of spatial and temporal message-passing dependencies, reducing redundant information exchange~\citep{zhang2025agentprune}. These methods adapt participating agents or communication structure. They provide structure-level comparisons to MAS-Gate's learned selection among both content-level and workflow-level interventions.

\subsubsection{Post-hoc attribution}


\paragraph{Direct LLM.}
\label{app:error_attribution_baseline}
Our baseline follows the All-at-Once formulation in Who\&When~\citep{zhang2025agent}, with the
reference task answer included in the input. Given the task query, complete trajectory, and reference answer, the localizer predicts an error step and its agent on Who\&When, or a single responsible agent on AEGIS, together with a brief explanation. MAS-Gate-guided inference uses the same localizer, task input, and output format, and appends the risk-based candidate ranking. 


\subsection{Implementation Details}
\label{app:implementation_details}

\subsubsection{Design-time orchestration}
\label{app:implementation_details_exp1}

Table~\ref{tab:implementation_details} summarizes key hyperparameters used in our reinforcement learning. Qwen3-8B~\citep{qwen3technicalreport} is the only trainable model and serves as the design-time Orchestra and runtime Editor. Qwen3-30B-A3B-Instruct-2507~\citep{qwen3technicalreport} remains frozen while executing sub-agent nodes and providing the teacher-forced likelihoods used by the within-node utility. The frozen bge-m3-zeroshot-v2.0~\citep{laurer_building_2023} service provides the entailment probability used for inter-node propagation. For ALFWorld, we construct the reference label $y$ as an abstract representation of the final state specified by the training example and use it for within-node utility quantification.

\subsubsection{Runtime intervention}
\label{app:implementation_details_exp2}

\textbf{Runtime Editor protocol.} We freeze the original Qwen3-8B checkpoint as the Orchestrator throughout Editor training and evaluation, independently of the orchestration training reported in Table~\ref{tab:main_results}. MAS-Gate supplies privileged training feedback through paired rewards evaluated on completed edited and unedited trajectories. Editor decisions are specified to condition on the task, graph, and observed execution prefix. At test time, the Editor receives neither the reference answer nor node-level quantification scores and makes at most one non-\textsc{Keep} intervention per execution.

\begin{table}[t]
\caption{Key hyperparameters used in our experiments.}
\label{tab:implementation_details}
\begin{center}
\small
\setlength{\tabcolsep}{4pt}
\begin{tabular}{p{0.43\linewidth}p{0.47\linewidth}}
\toprule
\textbf{Hyperparameter }& \textbf{Value} \\
\midrule
RL framework & verl \\
GPU & 8$\times$H100 \\
Train batch size & 256 \\
Rollout group size & 16 \\
PPO mini-batch size & 256 \\
PPO micro-batch size per GPU & 2 \\
Log-probability micro-batch size per GPU & 4 \\
Maximum prompt length & 15{,}000 \\
Maximum response length & 8{,}192 \\
Actor learning rate & $1\times10^{-6}$ \\
PPO epochs & 1 \\
PPO clipping ratio & 0.2 \\
KL coefficient & 0.001 \\
Gradient clipping norm & 1.0 \\
Maximum concurrency & 1{,}024\\
\bottomrule
\end{tabular}
\end{center}
\end{table}

\textbf{Runtime Editor fine-tuning.}
We optimize LoRA adapters for the Editor using GRPO, with rank 64, scaling factor $\alpha=32$, and zero dropout, applied to all linear layers while keeping the base-model weights frozen. Each episode allows at most four additional node executions. In Eq.~\ref{eq:runtime_reward}, we use $\beta=0.1$ and $\lambda=1.0$. Action costs are 0.03 for annotation and child guidance, 0.08 for retry, 0.10 for insertion, and 0.12 for backtracking; invalid actions receive a reward of $-0.25$.

\textbf{Action Space.}
The six actions operate on different parts of the execution. \textsc{Keep} preserves the current output and advances to the next eligible gate. \textsc{Retry-Current} re-executes the current node using its recorded prompt and context. \textsc{Backtrack-Parent} selects a direct parent and re-executes that parent together with its already executed affected descendants. \textsc{Annotate-Current} preserves the raw current output but appends a bounded diagnosis for downstream nodes. \textsc{Guide-Child} adds a bounded operational instruction to one unexecuted direct child without introducing a new node, whereas \textsc{Insert-Subtask} inserts one LLM-only verification, bridge, or supplement node between the current node and such a child. The legal set is constructed dynamically from the graph position, execution status, replay safety, and remaining node-call budget.

\subsubsection{Post-hoc attribution}
\label{app:implementation_details_exp3}

We use Qwen3-30B-A3B-Instruct-2507 for utility estimation and bge-m3-zeroshot-v2.0 for inter-node entailment scoring. Each recorded trajectory is represented as a chronological chain, with edges connecting consecutive steps as a proxy for temporal dependencies. The resulting MAS-Gate risk scores are computed once and reused across all localizers. The attribution results therefore characterize MAS-Gate under a temporal chain approximation.

The five localizers are Qwen3-30B-A3B-Instruct-2507, Qwen3.5-4B, Qwen3.6-27B, Mistral-Small-3.2-24B-Instruct-2506, and Llama-3.1-8B-Instruct. All localizers are run with vLLM using a temperature of $0$, a maximum output length of 256 tokens, and an 81,920-token context window. Thinking mode is disabled where supported. Direct LLM and MAS-Gate-guided inference use the same inference settings and retain the complete trajectory and reference task answer.

We set $k=8$ on Who\&When and $k=4$ on AEGIS. For AEGIS, steps are grouped by their recorded agent names, and each agent is ranked by its maximum eligible-step risk. User and System turns are retained as context but excluded from attribution targets. When fewer than $k$ candidates are available, all are included. Guidance consists of ranked step indices on Who\&When and ranked agent names on AEGIS; numerical risk scores are not included in the prompts. The prompt templates are provided in Appendix\ref{app:prompt_details}.

\section{Additional Ablation Experiments}
\label{app:additional_ablations}

\subsection{Design-time orchestration}
\label{app:additional_ablations_exp1}
The following ablations are organized around the principal claims made by the method rather than around arbitrary parameter sweeps. Each ablation changes one component while holding the training split, sub-agent backbone, total execution budget, and evaluation protocol fixed. Table~\ref{tab:design_ablation_results} reports the resulting accuracy on HotpotQA and the out-of-task BrowseComp-Plus benchmark.



\subsubsection{Alternative Node-Impact Composition}
\label{app:ablation_impact}

We compare the default multiplicative impact $I_i=U_i(1+K_i)$ with the additive alternative $I_i=U_i+(1+K_i)$. All other components, including the within-node utility, inter-node strength, training data, and evaluation protocol, remain unchanged. As shown in Table~\ref{tab:design_ablation_results}, the additive variant reaches 66.50 on HotpotQA and 11.90 on BrowseComp-Plus, below the corresponding MAS-Gate results of 69.50 and 13.69. We hypothesize that the multiplicative form better represents the intended interaction between node utility and propagation strength: downstream influence scales the value of the information produced by a node, so the two factors jointly determine its impact. In contrast, the additive form treats local utility and propagation more independently, allowing one component to offset the other and thereby weakening the distinction between information that is locally useful and information that is also effectively transmitted through the workflow.

\subsubsection{Design-Time Reward Ablation}
\label{app:ablation_reward}

This ablation tests whether the design-time Orchestra benefits from the
node-centric dense signal beyond terminal correctness. We train two otherwise
identical Orchestra policies with the same initialization, rollout groups,
optimization budget, sub-agent backbone, and task mixtures. The first policy
uses only the terminal accuracy signal,
\begin{equation}
    r_{\mathrm{acc}}(\tau)=c(\tau),
\end{equation}
where $c(\tau)\in\{0,1\}$ indicates whether the completed execution produces a
correct answer. The second policy uses the composite reward from the main
experiment,
\begin{equation}
    r_{\mathrm{dense+acc}}(\tau)=
    \frac{1}{|V|}\sum_{i\in V} I_i+c(\tau).
\end{equation}
Thus, the comparison changes only the training signal: both policies are
prompted to generate the same class of executable workflows and are evaluated
with the same terminal accuracy metric. Table~\ref{tab:design_ablation_results}
shows that the fused reward improves HotpotQA accuracy from 64.50 to 69.50 and
BrowseComp-Plus accuracy from 9.52 to 13.69, corresponding to gains of 5.00 and
4.17 percentage points, respectively, or 4.59 points on average. The
out-of-task result is particularly informative: accuracy-only training falls
below the CoT baseline on BrowseComp-Plus (9.52 versus 11.31 in
Table~\ref{tab:main_results}), whereas the fused reward surpasses it with
13.69. This reversal suggests that sparse binary correctness can overfit the
source-task outcome without adequately distinguishing the quality of
intermediate coordination. By incorporating graded node-level contributions,
the fused reward provides a richer training signal and yields stronger
transfer to workflows required by the unseen task distribution.

\begin{table}[t]
\caption{Ablation results for node-impact composition and design-time reward. BCP denotes the out-of-task BrowseComp-Plus evaluation. All values are accuracy (\%).}
\label{tab:design_ablation_results}
\begin{center}
\begin{tabular}{lcc}
\toprule
Training variant & HotpotQA & BCP\textsubscript{OOT} \\
\midrule
Full MAS-Gate & \textbf{69.50} & \textbf{13.69} \\
Additive impact & 66.50 & 11.90 \\
Accuracy-only reward & 64.50 & 9.52 \\
\bottomrule
\end{tabular}
\end{center}
\end{table}

\subsection{Post-hoc attribution}
\label{app:additional_ablations_exp3}

\subsubsection{Error Attribution Guidance Ablation}
\label{app:error_attribution_ablation}

This ablation examines the contributions of candidate selection and
risk-based ordering to error localization. We use Qwen3-30B-A3B on
Who\&When with $k=8$, keeping the complete trajectory, reference task
answer, and inference settings unchanged. In addition to Direct LLM,
we compare three guided variants: randomly sampled candidates,
MAS-Gate candidates presented in shuffled order, and MAS-Gate
candidates ranked by risk. The guided variants use the same prompt
template and differ only in candidate selection or presentation order.

Table~\ref{tab:attribution_ablation} shows that random
candidate guidance achieves 34.78\% accuracy, compared with 33.70\%
for Direct LLM. Shuffled MAS-Gate candidates achieve 35.33\%,
while retaining their risk-based ranking yields 37.50\%.
Since the ranked and shuffled variants use the same candidate
steps, this comparison provides evidence that the ordering itself
offers useful information for localization. Together, these results
suggest that the localizer benefits from both MAS-Gate's selection
of suspicious steps and its ranking of these steps by risk.

\begin{table}[t]
\caption{Ablation of risk-based guidance on Who\&When with
Qwen3-30B-A3B as the localizer and $k=8$.
Random Top-$8$ provides randomly selected candidate steps,
whereas MAS-Gate (shuffled) presents the MAS-Gate candidates
in random order. We report exact step-level accuracy (\%).}
\label{tab:attribution_ablation}
\begin{center}
\setlength{\tabcolsep}{8pt}
\begin{tabular}{lc}
\toprule
Method & Accuracy (\%) \\
\midrule
Direct LLM & 33.70 \\
+ Random Top-$8$ & 34.78 \\
+ MAS-Gate (shuffled) & 35.33 \\
+ MAS-Gate & \textbf{37.50} \\
\bottomrule
\end{tabular}
\end{center}
\end{table}

\section{Case Study}
\label{app:case_study}

\subsection{Design-Time Orchestration}
\label{app:case_design_time}

We select one HotpotQA instance for which an untrained Orchestra generates an MAS that produces an incorrect answer, whereas the Orchestra trained with MAS-Gate produces the correct answer (i.e., Skill-MAS vs. MAS-Gate). In this instance, Skill-MAS returns \texttt{None}, whereas MAS-Gate correctly returns \textbf{Martin O'Neill}. Skill-MAS spends four of five nodes repeatedly resolving or verifying the easier bridge entity, leaving the harder successor relation to one terminal node. Contrastively, trained MAS-Gate resolves the bridge once, then allocates dedicated nodes to successor discovery, timeline verification, and answer synthesis.

\input{sections/case_study}

\subsection{Runtime Intervention}
\label{app:case_runtime}


We illustrate runtime intervention on a HotpotQA instance using the same initial three-node workflow and observed prefix. Without intervention, the downstream continuation returns an incorrect financing provider. A single \textsc{Retry-Current} action produces a more informative retrieval output, allowing the unchanged downstream workflow to answer correctly under semantic evaluation.

\begin{caseoverview}{Case information}
\textbf{Benchmark.} HotpotQA, task ID \texttt{5ab5cd0d554299488d4d9a30}.\\
\textbf{Question.} ``The not-for-profit company which owns a Mark VI monorail was assisted in bond financing by whom?''\\
\textbf{Gold answer.} The State of Nevada.\\
\textbf{No-intervention answer.} Federal Transit Administration (FTA).\\
\textbf{Edited answer.} The State of Nevada provided bond financing assistance to the Las Vegas Monorail Company.\\
\textbf{Evaluation.} Semantic correctness changes from $0$ to $1$; this is not a string-level exact-match comparison.
\end{caseoverview}

\begin{tracebox}{Workflow and observed handoff}
The workflow contains three nodes: \texttt{agent\_0} retrieves the company using the BM25 corpus, \texttt{agent\_1} identifies its bond-financing provider from the retrieved information, and \texttt{agent\_2} synthesizes the answer. The initial output of \texttt{agent\_0} includes:
\begin{quote}\small\ttfamily
The Las Vegas Monorail Company is a not-for-profit corporation that owns and operates the Las Vegas Monorail, which uses Mark VI monorail trains.
\end{quote}
The company is correctly identified, but the handoff does not supply the financing provider required downstream. The Editor is queried at this point, before either downstream node executes.
\end{tracebox}

\begin{tracebox}{Recorded Editor action}
The Editor emits:
\begin{quote}\small\ttfamily
\{"action":"RETRY\_CURRENT","target":"agent\_0","payload":\{\}\}
\end{quote}
The runtime re-executes the current node with its saved prompt and context. The graph remains unchanged. The action record contains no explicit natural-language diagnosis and does not provide a replacement answer.
\end{tracebox}

\begin{failurebox}{No-intervention continuation}
With the original handoff, \texttt{agent\_1} produces:
\begin{quote}\small\ttfamily
The entity that provided bond financing assistance to the Las Vegas Monorail Company was the Federal Transit Administration (FTA).
\end{quote}
The final node repeats \texttt{Federal Transit Administration (FTA)}. This incorrect entity appears after the intervention point and was not visible to the Editor when it selected its action.
\end{failurebox}

\begin{successbox}{Edited continuation}
After the retry, \texttt{agent\_0} supplies the missing financing relation:
\begin{quote}\small\ttfamily
The State of Nevada provided bond financing assistance for the project, though no public funds were used in construction.
\end{quote}
Both downstream nodes subsequently identify the State of Nevada as the financing provider. The final answer semantically matches the gold answer. This example illustrates how a targeted retry can improve the information available to downstream agents without regenerating the workflow or having the Editor directly write the answer.
\end{successbox}

\subsection{Post-hoc Attribution}
\label{app:case_posthoc}

We illustrate post-hoc attribution on a Who\&When trajectory in which
incorrect sale-price information is repeatedly accepted by downstream
agents. Given the same complete trajectory and reference task answer,
Direct LLM attributes the failure to a later validation step, whereas
the MAS-Gate-guided localizer identifies the annotated decisive error
where the incorrect information first enters the trajectory.

\begin{caseoverview}{Case information}
\textbf{Benchmark.} Who\&When, trajectory
\texttt{Algorithm-Generated/4}.\\
\textbf{Task.} Compare the 2022 sale prices of 2072 Akaikai Loop and
2017 Komo Mai Drive in Pearl City, Hawaii, and return the higher price
without commas or decimal places.\\
\textbf{Reference answer.} \texttt{900000}.\\
\textbf{Recorded final answer.} \texttt{950000}.\\
\textbf{Annotated decisive error.} Step~2,
\texttt{HawaiiRealEstate\_Expert}.\\
\textbf{Localizer.} Qwen3-30B-A3B, with $k=8$ for MAS-Gate guidance.\\
\textbf{Indexing and evaluation.} Steps are zero-based.
Direct LLM selects Step~4; MAS-Gate-guided attribution selects Step~2,
matching the released annotation. Both predictions concern the same
failed execution.
\end{caseoverview}

\begin{tracebox}{Introduction and subsequent repetition of the error}
At Step~2, \texttt{HawaiiRealEstate\_Expert} first supplies the sale
prices. The recorded price claims are excerpted below, with formatting
simplified:
\begin{quote}\small\ttfamily
2072 Akaikai Loop:\\
Sold for \$850,000\\[2pt]
2017 Komo Mai Drive:\\
Sold for \$950,000
\end{quote}
At Steps~3--4, \texttt{Validation\_Expert} repeats these prices,
compares them, and accepts \texttt{950000} as the answer.
Its Step~4 response concludes:
\begin{quote}\small\ttfamily
Since all validation checks are met and the conclusion aligns with
the provided data, we can consider this task complete.
\end{quote}
At Step~6, \texttt{Data\_Analysis\_Expert} again reports
\texttt{950000}. The released annotation assigns the decisive error
to Step~2, explaining that the incorrect price propagates through
subsequent steps to the final output.
\end{tracebox}

\begin{failurebox}{Direct LLM attribution}
Direct LLM selects \textbf{Step~4},
\texttt{Validation\_Expert}, and gives the following explanation:
\begin{quote}\small\ttfamily
Incorrectly accepted and reported \$950,000 as the sale price without
verifying against the correct answer of 900000, indicating a failure
to validate against authoritative data.
\end{quote}
This diagnosis focuses on the downstream acceptance of the incorrect
price, rather than the earlier step identified by the annotation.
\end{failurebox}

\begin{successbox}{MAS-Gate-guided attribution}
MAS-Gate assigns its highest risk score to \textbf{Step~2}.
This trajectory contains eight recorded steps, so the top-$8$ guidance
includes every step and adds their risk-based ordering to the prompt.
With this guidance, the same localizer selects \textbf{Step~2},
\texttt{HawaiiRealEstate\_Expert}, and explains:
\begin{quote}\small\ttfamily
Provided incorrect sale prices for the homes, stating 2072 Akaikai
Loop sold for \$850,000 and 2017 Komo Mai Drive for \$950,000, when
the correct answer is 900000 for the higher-priced home.
\end{quote}
The predicted agent and step both match the released annotation.
This case illustrates how MAS-Gate guidance can help the localizer
identify where erroneous information first enters a trajectory,
even when subsequent agents repeatedly endorse it.
\end{successbox}

\section{Prompt Details}
\label{app:prompt_details}

Figures~\ref{fig:mas-gate-xml-orchestrator-prompt}--\ref{fig:llm_judge_vita} present the five prompts used in our experiments: the Orchestrator prompt in Figure~\ref{fig:mas-gate-xml-orchestrator-prompt}, the Editor prompt in Figure~\ref{fig:mas-gate-editor-prompt}, the LLM-based error-attribution prompt for Experiment~3 from Figure~\ref{fig:who_when_attribution_prompt} to Figure~\ref{fig:aegis_attribution_prompt}, and the VitaBench LLM-as-a-judge prompt in Figure~\ref{fig:llm_judge_vita}.

\begin{figure*}[htbp]
    \centering
    \begin{tcolorbox}[
        enhanced,
        colframe=Salmon!90!Black,
        colback=Salmon!20,
        coltitle=white,
        fonttitle=\large\bfseries,
        title={MAS-Gate Prompt for Orchestrator},
        halign title=left,
        fontupper=\footnotesize,
        boxrule=1pt,
        arc=3mm,
        boxsep=2pt,
        left=8pt,
        right=8pt,
        top=4pt,
        bottom=4pt
    ]

    \textbf{System\_prompt:}

    You are MAS-Gate's orchestra policy. Design a multi-agent system, not the final answer.

    Return XML only. Do not solve the problem directly.

    \textbf{User\_prompt:}

    Dataset: \{dataset\_name\}

    Task:

    \{question\}

    Output exactly one XML document with this shape:\\

    \texttt{<editor\_output>}\\
    \hspace*{2em}\texttt{<task\_decomposition>}\\
    \hspace*{4em}\texttt{<goal>overall objective</goal>}\\
    \hspace*{4em}\texttt{<sub\_tasks>}\\
    \hspace*{6em}\texttt{<sub\_task idx="0">...</sub\_task>}\\
    \hspace*{6em}\texttt{<sub\_task idx="1">...</sub\_task>}\\
    \hspace*{4em}\texttt{</sub\_tasks>}\\
    \hspace*{2em}\texttt{</task\_decomposition>}\\
    \hspace*{2em}\texttt{<agent\_specs>}\\
    \hspace*{4em}\texttt{<agents>}\\
    \hspace*{6em}\texttt{<agent id="agent\_0" is\_final="false" execution\_mode="llm\_only">}\\
    \hspace*{8em}\texttt{<role>specialized role and instructions</role>}\\
    \hspace*{8em}\texttt{<sub\_task\_idx>0</sub\_task\_idx>}\\
    \hspace*{8em}\texttt{<dependencies></dependencies>}\\
    \hspace*{6em}\texttt{</agent>}\\
    \hspace*{4em}\texttt{</agents>}\\
    \hspace*{2em}\texttt{</agent\_specs>}\\
    \texttt{</editor\_output>}\\

    \end{tcolorbox}

    \caption{MAS-Gate XML Orchestrator prompt. The user prompt is assembled dynamically from the task, dataset name, and dataset-specific routing/output constraints.}
    \label{fig:mas-gate-xml-orchestrator-prompt}
\end{figure*}




\begin{figure*}[p]
\centering
\begin{tcolorbox}[
    enhanced,
    colframe=Salmon!90!Black,
    colback=Salmon!20,
    coltitle=white,
    fonttitle=\large\bfseries,
    title={MAS-Gate Prompt for Editor},
    halign title=left,
    fontupper=\footnotesize\raggedright,
    boxrule=1pt, arc=3mm, boxsep=2pt,
    left=8pt, right=8pt, top=4pt, bottom=4pt
]
\textbf{System prompt (excerpt):}

You are the runtime editor for a multi-agent system.
At each gate, inspect only the visible graph prefix and select one legal action.

\textbf{Decision precedence:}

1. First classify whether the current milestone completed its visible role, task, constraints, and (for a final node) the requested global deliverable.

2. If it is clearly INCOMPLETE because the current node omitted or failed its assigned deliverable, \texttt{RETRY\_CURRENT} has positive expected value and you must select \texttt{RETRY\_CURRENT} when it is legal; \texttt{KEEP} is not valid for that case.

3. If the visible failure instead originates in a parent handoff, select \texttt{BACKTRACK\_PARENT} when legal.

4. Only after the completion check, compare disruption and extra-call costs among suitable repairs. For a COMPLETE milestone, \texttt{KEEP} unless a specific legal repair has positive expected net value. Do not invent an error to justify editing.

\textit{[Additional instructions omitted.]}

Do not solve the global task yourself. Return exactly one JSON object matching:

\texttt{\{"action":"ACTION","target":"NODE\_OR\_NULL","payload":\{...\}\}}

\texttt{KEEP} has null target and an empty payload. Use only the legal actions and targets shown in the observation. Never invent hidden outputs or a ground-truth answer.

\medskip
\textbf{User prompt (excerpts):}

\textit{[Milestone-completion checks and the action-selection guide omitted.]}

\textbf{Payload rules:}

\texttt{ANNOTATE\_CURRENT}: \texttt{error\_type}, \texttt{target\_span}, \texttt{diagnosis}, \texttt{downstream\_check}. \texttt{error\_type} must be one of LOGIC, CALCULATION, EVIDENCE, CONSTRAINT, INCOMPLETE, CONFLICT, LOW\_CONFIDENCE. Copy \texttt{target\_span} only from \texttt{observed\_outputs[current\_node]}, never from the question, role, or task. Keep it under 12 words and never write a corrected answer. Maximum lengths are 24 tokens for \texttt{target\_span}, 48 for \texttt{diagnosis}, 24 for \texttt{downstream\_check}, and 96 total.

\texttt{GUIDE\_CHILD}: \texttt{guidance\_type} and \texttt{focus}. \texttt{guidance\_type} is one of VERIFY\_UPSTREAM, RECOMPUTE, CHECK\_CONSTRAINT, RESOLVE\_CONFLICT. \texttt{focus} is at most 32 tokens.

\texttt{INSERT\_SUBTASK}: \texttt{mode} must be VERIFY, BRIDGE, or SUPPLEMENT. \texttt{task} is at most 48 tokens and is required for BRIDGE and SUPPLEMENT.

\texttt{KEEP}, \texttt{RETRY\_CURRENT}, and \texttt{BACKTRACK\_PARENT} use an empty payload.

\textit{[JSON examples omitted.]}

OBSERVATION: \texttt{\{public\_observation\_json\}}

\medskip
\textbf{Training follow-up after KEEP:}

Next gate. Apply the same decision rules to this new observation.

OBSERVATION: \texttt{\{next\_gate\_observation\_json\}}
\end{tcolorbox}
\caption{Excerpts from the Runtime Editor prompt. Observations are populated dynamically with public execution information and action constraints. The final block shows the training-time follow-up after \textsc{Keep}.}
\label{fig:mas-gate-editor-prompt}
\end{figure*}

\begin{figure*}[htbp]
\centering
\begin{tcolorbox}[
enhanced,
colframe=Salmon!90!Black,
colback=Salmon!20,
coltitle=white,
fonttitle=\large\bfseries,
title={LLM-Based Error Attribution Prompt (Who\&When)},
halign title=left,
fontupper=\footnotesize,
boxrule=1pt,
arc=3mm,
boxsep=2pt,
left=8pt,
right=8pt,
top=4pt,
bottom=4pt
]
\textbf{System\_prompt:}

You are a precise failure-attribution evaluator. Follow the requested output
format exactly and do not add headings, markdown fences, or extra lines.

\medskip
\textbf{User\_prompt:}

Analyze the failed multi-agent trajectory below.\\
Problem: \{question\}\\
Reference correct answer: \{reference\_answer\}\\
Valid Step Number range: 0 to \{last\_step\}, inclusive.\\
Identify the single agent directly responsible for the wrong solution and
the first zero-based step where that agent made the decisive mistake.
Agent Name must exactly match the Agent field shown at the selected step.
Give a concise one-line reason.\\
Return exactly these three lines:\\
\texttt{Agent Name: <name>}\\
\texttt{Step Number: <integer>}\\
\texttt{Reason for Mistake: <reason>}

Conversation:\\
\{complete\_trajectory\}

\medskip
\textbf{Additional user-prompt suffix (MAS-Gate only):}

Additional MAS-Gate attribution information:\\
MAS-Gate is a computation-based failure-attribution method that scores and
ranks trajectory steps using their utility and communication influence.
Its ranking provides useful information for identifying the decisive error.
The following steps received the highest scores and are therefore the most
suspicious candidates.

MAS-Gate candidate ranking:\\
Rank 1: Step \{step\_1\}\\
\ldots\\
Rank \{effective\_k\}: Step \{step\_effective\_k\}

Use the MAS-Gate ranking as an important reference when identifying the
single agent and the first decisive-error step.

\end{tcolorbox}

\caption{LLM prompts for step-level error attribution on Who\&When.
Direct LLM uses the system and user prompts. MAS-Gate additionally
appends the ranked-step guidance after the complete trajectory.
Braced fields are replaced with sample-specific content, and the
ranking contains up to $k$ entries.}
\label{fig:who_when_attribution_prompt}
\end{figure*}

\begin{figure*}[htbp]
\centering
\begin{tcolorbox}[
enhanced,
colframe=Salmon!90!Black,
colback=Salmon!20,
coltitle=white,
fonttitle=\large\bfseries,
title={LLM-Based Error Attribution Prompt (AEGIS)},
halign title=left,
fontupper=\footnotesize,
boxrule=1pt,
arc=3mm,
boxsep=2pt,
left=8pt,
right=8pt,
top=4pt,
bottom=4pt
]
\textbf{System\_prompt:}

You are a precise failure-attribution evaluator. Follow the requested
two-line output format exactly and do not add headings or markdown.

\medskip
\textbf{User\_prompt:}

Analyze the failed multi-agent trajectory below.\\
Problem: \{question\}\\
Reference correct answer: \{reference\_answer\}\\
Observed final output: \{observed\_final\_output\}\\
Valid selectable agents: \{valid\_agent\_names\}.\\
User and System turns are context only and cannot be selected. Choose
exactly one valid agent. Agent Name must exactly match one name in the
valid selectable agents list.\\
Return exactly these two lines:\\
\texttt{Agent Name: <name>}\\
\texttt{Reason for Mistake: <one-line reason>}

Conversation:\\
\{complete\_trajectory\}

\medskip
\textbf{Additional user-prompt suffix (MAS-Gate only):}

MAS-Gate identified the following Top-\{effective\_k\} most suspicious
agents, ranked by estimated error risk from highest to lowest.
Rank 1 is the most suspicious:\\
Rank 1: \{agent\_1\}\\
\ldots\\
Rank \{effective\_k\}: \{agent\_effective\_k\}\\
Use this ranking together with the complete trajectory and the
reference correct answer to identify the agent responsible for
the failure.

\end{tcolorbox}

\caption{LLM prompts for agent-level error attribution on AEGIS.
Direct LLM uses the system and user prompts. MAS-Gate additionally
appends agents ranked by their maximum eligible-step risk after
the complete trajectory. Braced fields are replaced with
sample-specific content, and the ranking contains up to $k$ entries.}
\label{fig:aegis_attribution_prompt}
\end{figure*}

\begin{figure*}[htbp]
    \centering
    \begin{tcolorbox}[
        enhanced,
        colframe=Salmon!90!Black,
        colback=Salmon!20,
        coltitle=white,
        fonttitle=\large\bfseries,
        title={LLM-as-a-judge Prompts (VitaBench)},
        halign title=left,
        fontupper=\footnotesize,
        boxrule=1pt,
        arc=3mm,
        boxsep=2pt,
        left=8pt,
        right=8pt,
        top=4pt,
        bottom=4pt
    ]

\hspace{0em}\textbf{System Prompt:}
System Information
\texttt{\{env\_info\}}
User Complete Instruction
\texttt{\{user\_instruction\}}

\textbf{Background}
- This is a conversation scenario between a user and an assistant, where the assistant can call tools to retrieve information and complete operations. Tool return results will start with "tool"

\hspace{4em}- You need to evaluate whether the user instruction has been completed. The user's complete instruction has been broken down into several scoring points (rubrics), and you only need to judge whether each scoring point is satisfied

\hspace{4em}- Due to the large number of conversation turns, we use sliding window evaluation, where each window shows 10 conversation turns with 2 overlapping turns between windows, but rubric status will be preserved across windows

\hspace{4em}- You are evaluating window \texttt{\{window\_idx\}} (out of \texttt{\{total\_windows\}} windows total)

\hspace{4em}- \texttt{<window\_content>} contains the conversation content for the current window

\hspace{4em}- \texttt{<current\_rubrics>} contains the current status of all scoring points (true means satisfied, false means not satisfied, all scoring points have an initial status of false)

\hspace{2em}\textbf{Task}
- Update the scoring point rubric status based on the conversation content in the current window

\hspace{4em}- You can update the status from false to true, if and only if the assistant completed the goal in this window

\hspace{4em}- You can also update true back to false, if and only if the assistant overturned a previous correct conclusion in this window (but note: if the user's needs themselves changed, such as actively canceling after placing an order, the original evaluation of the ordering requirement should not be overturned)

\hspace{4em}- You can refer to the "User Complete Instruction" to understand the progress of the current conversation window and avoid unnecessary modifications

\hspace{2em}\textbf{Important Notes}

\hspace{4em}- Important: All evaluations are based on whether the assistant's responses and tool call requests complete the goals in the rubrics. User expressions in the conversation are only considered as guidance for the assistant and do not directly affect evaluation standards. Everything is based on the rubric fields!

\hspace{4em}- Important: Query tool return results are only visible to the assistant and do not represent content recommended by the assistant to users, so they do not directly affect evaluation results either. Everything must be based on the assistant's responses to users after obtaining information! Also note that the Assistant cannot fabricate Tool return results!

\hspace{4em}- Important: For order-related rubrics (involving order details that must generate orders), you must confirm whether the assistant actually completed the ordering operation. The assistant may mistakenly believe they completed the ordering operation when in fact the tool call failed; or situations where the user states they "can place the order themselves," etc., should all be considered as not meeting the requirements

\hspace{4em}- For rubrics involving order details such as product quantity or delivery time, the original rubric requirements must be strictly met (no deviation in product quantities, delivery must not be later than the expected time). User compromise behavior does not affect evaluation results (for example, when a user states "fewer items is okay", "I have no objections to the order content" or "later delivery is fine" etc.). These situations should still be considered as not meeting the requirements

\hspace{4em}- For rubrics involving text content matching of addresses or order notes, apply the functional equivalence principle: as long as the actual content can achieve the same function (such as roughly locating the delivery location or conveying the customer's main needs), it is considered to meet the requirements even if the expression is not completely consistent or lacks some details

\hspace{4em}- If the current window does not involve a certain rubric, keep its original status unchanged; if the current window involves a rubric but cannot fully determine it, key information can be recorded in the justification for future judgment

\hspace{4em}- In the justification, record key information related to the current rubric and its corresponding round [x] in an appending manner. If status modifications occur, record the reason using concise language. If the status is not modified, restate the previous reason

\hspace{2em}Format Requirements

\hspace{4em}- Your response should be a JSON object containing the following fields:

\hspace{6em}- \texttt{rubric\_idx}: Unique identifier for the rubric

\hspace{6em}- \texttt{rubric}: Restatement of the rubric

\hspace{6em}- \texttt{justification}: Explanation of status changes

\hspace{6em}- \texttt{meetExpectation}: Updated status (true or false)

\hspace{0em}\textbf{User Prompt:}
\hspace{2em}\texttt{<window\_content>}
\texttt{\{window\_content\}}
\hspace{0em}\texttt{</window\_content>}

\hspace{2em}\texttt{<current\_rubrics>}
\texttt{\{current\_rubrics\_str\}}
\hspace{0em}\texttt{</current\_rubrics>}

    \end{tcolorbox}
    \caption{LLM-as-a-judge prompts used in VitaBench.}
    \label{fig:llm_judge_vita}
\end{figure*}

%% file: sections/case_study.tex
\begin{caseoverview}{Case information}
\textbf{Benchmark.} HotpotQA, task ID \texttt{5ac175635542994ab5c67d4b}.\\
\textbf{Question.} ``Who replaced the manager of the Aston Villa that began at Leeds United?''\\
\textbf{Label.} \textbf{Martin O'Neill}.\\
\textbf{Skill-MAS's Answer.} None. \textcolor{red}{$\times$}\\
\textbf{MAS-Gate's Answer.} Martin O'Neill. \textcolor{blue}{$\checkmark$}
\end{caseoverview}

\begin{tracebox}{Skill-MAS Node 1 --- \texttt{extract\_entities}}
\textbf{Role.} Extract manager and club candidates explicitly mentioned in retrieved evidence.\\
\textbf{Retrieved/used evidence.} \emph{David O'Leary}, \emph{Mervyn Day}, \emph{Ian Culverhouse}, \emph{G\'erard Houllier}, and Aston Villa season pages.\\
\textbf{Response.}
\begin{quote}\small\ttfamily
managers = [David O'Leary, Ian Culverhouse, G\'erard Houllier, Tim Sherwood, R\'emi Garde, Eric Black]\\
clubs = [Aston Villa, Leeds United, Al-Ahli Dubai, West Ham United,
Leyton Orient, Luton Town, Sheffield United, Carlisle United]
\end{quote}
\end{tracebox}

\begin{tracebox}{Skill-MAS Node 2 --- \texttt{find\_leeds\_manager}}
\textbf{Role.} Determine which extracted manager began a managerial career at Leeds United.\\
\textbf{Direct predecessor.} \texttt{extract\_entities}.\\
\textbf{Retrieved/used evidence.} \emph{David O'Leary}: managerial career began at Leeds United; \emph{Mervyn Day} and \emph{1996 Football League Cup Final} act as lexical distractors.\\
\textbf{Response.}
\begin{quote}\small\ttfamily David O'Leary \end{quote}
\end{tracebox}

\begin{tracebox}{Skill-MAS Node 3 --- \texttt{find\_villa\_manager}}
\textbf{Role.} Check that Leeds-linked candidate also managed Aston Villa.\\
\textbf{Direct predecessor.} \texttt{find\_leeds\_manager}.\\
\textbf{Retrieved/used evidence.} \emph{David O'Leary} states that he later managed Aston Villa; \emph{List of Aston Villa F.C. managers} provides club-manager context.\\
\textbf{Response.}
\begin{quote}\small\ttfamily David O'Leary \end{quote}
\end{tracebox}

\begin{tracebox}{Skill-MAS Node 4 --- \texttt{verify\_relation}}
\textbf{Role.} Verify relation between identified Aston Villa manager and Leeds United.\\
\textbf{Direct predecessor.} \texttt{find\_villa\_manager}.\\
\textbf{Retrieved/used evidence.} David O'Leary biography plus Aston Villa manager chronology.\\
\textbf{Response.}
\begin{quote}\small\ttfamily
David O'Leary was the manager of Aston Villa and had a prior association with Leeds United.
\end{quote}
\end{tracebox}

\begin{failurebox}{Skill-MAS Node 5 --- \texttt{identify\_replacement} (localized failure)}
\textbf{Role.} Identify who replaced the resolved Aston Villa manager and return one supported name.\\
\textbf{Direct predecessor.} \texttt{verify\_relation}. \\
\textbf{Retrieved/used evidence.} Node 5 does receive two relevant pieces of evidence: (i) \emph{David O'Leary} states that his managerial career began at Leeds United and that he later managed Aston Villa; and (ii) \emph{2010--11 Aston Villa F.C. season} identifies Martin O'Neill as Aston Villa's previous manager before G\'erard Houllier. Thus, candidate \textbf{Martin O'Neill} is already present in the evidence. Still, the relation between the two pieces is indirect: retrieved snippets do not explicitly state that Martin O'Neill replaced David O'Leary. \\
\textbf{Expectation.} Resolve exact successor predicate $\textsc{ReplacedAt}(x,\text{David O'Leary},\text{Aston Villa})$, retrieve or cross-check direct chronology evidence, and emit one name.\\
\textbf{Response.}
\begin{quote}\small\ttfamily
Based on the verified relation that David O'Leary was the manager of Aston Villa and had a prior association with Leeds United, the evidence does not explicitly provide the name of the person who replaced him at Aston Villa. Therefore, no supported candidate can be determined from the given evidence.\\
Final answer: None
\end{quote}
\textbf{Failure cause.} Failure is best characterized as \textbf{evidence integration and completion failure}, not complete absence of relevant evidence. Node 5 sees David O'Leary bridge fact and Martin O'Neill timeline cue, but fails to connect them into candidate chain \texttt{David O'Leary $\rightarrow$ Martin O'Neill} and then verify missing replacement edge. Once evidence is indirect, the node has no recovery stage for focused retrieval or independent verification and immediately returns \texttt{None}. MAS-Gate avoids this failure because \textbf{Successor Finder} uses the resolved David O'Leary entity to retrieve explicit replacement documents, while \textbf{Timeline Verifier} separately confirms the same edge.
\end{failurebox}

\begin{successbox}{MAS-Gate Node 1 --- Bridge Resolver (\texttt{agent\_0})}
\textbf{Role.} Resolve the Aston Villa manager described through Leeds United.\\
\textbf{Retrieved/used evidence.} The decisive document is \emph{David O'Leary} (\texttt{e4b9...ef5}), which states that his managerial career began at Leeds United and that he later managed Aston Villa. Retrieved distractors include \emph{Mervyn Day}, \emph{1996 Football League Cup Final}, \emph{List of Aston Villa F.C. managers}, \emph{Andy Marshall}, \emph{Joe Mercer}, \emph{Stiliyan Petrov}, and \emph{G\'erard Houllier}. \\
\textbf{Response.}
\begin{quote}\small\ttfamily
The Aston Villa manager who previously played at Leeds United is David O'Leary.
\end{quote}
\end{successbox}

\begin{successbox}{MAS-Gate Node 2 --- Successor Finder (\texttt{agent\_1})}
\textbf{Role.} Find replacement for the identified Aston Villa manager.\\
\textbf{Direct predecessor.} \texttt{agent\_0}; upstream output fixes bridge entity as \textbf{David O'Leary}.\\
\textbf{Retrieved/used evidence.} First-round retrieval returns \emph{2008--09 Aston Villa F.C. season} (\texttt{0544...50a}), which states that Martin O'Neill was in his third season since replacing David O'Leary, and \emph{2009--10 Aston Villa F.C. season} (\texttt{b597...79f}), which states that he was in his fourth season since replacing David O'Leary. \emph{David O'Leary} confirms bridge entity, while \emph{List of Aston Villa F.C. managers} provides chronology background.\\
\textbf{Response.}
\begin{quote}\small\ttfamily
The replacement of David O'Leary as the manager of Aston Villa was Martin O'Neill.
\end{quote}
\end{successbox}

\begin{successbox}{MAS-Gate Node 3 --- Timeline Verifier (\texttt{agent\_2})}
\textbf{Role.} Verify proposed replacement identity and temporal relation.\\
\textbf{Direct predecessor.} \texttt{agent\_1}; upstream candidate is \textbf{Martin O'Neill}, linked to predecessor \textbf{David O'Leary}.\\
\textbf{Retrieved/used evidence.} Retrieved titles include \emph{David O'Leary}, \emph{2008--09 Aston Villa F.C. season}, \emph{Mervyn Day}, \emph{O'Leary Manager 2000}, and \emph{1996 Football League Cup Final}. Explicit season evidence directly confirms that Martin O'Neill replaced David O'Leary.\\
\textbf{Response.}
\begin{quote}\small\ttfamily
David O'Leary was the manager of Aston Villa before being replaced. According to the evidence from ``2008--09 Aston Villa F.C. season,'' Martin O'Neill replaced David O'Leary as the manager of Aston Villa.\\
Thus, the manager who replaced David O'Leary at Aston Villa was Martin O'Neill.
\end{quote}
\end{successbox}

\begin{successbox}{MAS-Gate Node 4 --- Answer Synthesizer (\texttt{agent\_3})}
\textbf{Role.} Compose final answer from verified two-hop chain.\\
\textbf{Direct predecessor.} \texttt{agent\_2}; upstream output explicitly verifies \textbf{Martin O'Neill} as replacement of \textbf{David O'Leary}.\\
\textbf{Retrieved/used evidence.} No additional retrieval is needed because both required predicates have been closed upstream.\\
\textbf{Response.}
\begin{quote}\small\ttfamily
The manager of Aston Villa who previously played at Leeds United was David O'Leary. He was replaced by Martin O'Neill.
\end{quote}
\end{successbox}